%% file: main.tex
\documentclass[acmsmall]{acmart}

\usepackage{etoolbox}
\usepackage{enumitem}
\usepackage{comment}
\usepackage{xspace}
\usepackage{xr-hyper}
\usepackage{url}
\usepackage{graphicx}
\usepackage{balance}
\usepackage{xcolor}
\usepackage{amsmath}
\usepackage{enumitem}
\usepackage{subfigure}
\usepackage{cleveref}
\usepackage{multirow}
\usepackage{float}
\usepackage{framed}
\usepackage{listings}
\usepackage{algorithm}
\usepackage{algorithmicx}
\usepackage{algpseudocode}
\usepackage{booktabs}
\usepackage{dblfloatfix}  
\usepackage{CJK}
\usepackage{xcolor}
\usepackage{tikz}
\usepackage{listings}
\usepackage{subcaption}
\usepackage{array} 
\usetikzlibrary{arrows.meta,positioning,fit,calc,shapes.misc,shapes.symbols}
\usetikzlibrary{arrows.meta,positioning,fit,calc,shapes.misc,matrix,backgrounds}
\usepackage{pifont}
\usepackage{xcolor}
\definecolor{tablecolor}{RGB}{230,240,255}
\definecolor{correctcolor}{RGB}{220,255,220}
\definecolor{wrongcolor}{RGB}{255,230,230}

\definecolor{annotationcolor}{rgb}{0.8,0.0,0.0} 

\input{_commands}

\input{_notations}

\title{Reward-SQL: Boosting Text-to-SQL via Stepwise Execution-Aware Reasoning and Process-Supervised Rewards}

\author{Yuxin Zhang}
\email{yuxin.zhang@ruc.edu.cn}
\affiliation{%
  \institution{Renmin University of China}
  \city{Beijing}
  \country{China}
}

\author{Meihao Fan}
\email{fmh1art@ruc.edu.cn}
\affiliation{%
  \institution{Renmin University of China}
  \city{Beijing}
  \country{China}
}

\author{Ju Fan}
\authornote{Ju Fan is the corresponding author.}
\email{fanj@ruc.edu.cn}
\affiliation{%
  \institution{Renmin University of China}
  \city{Beijing}
  \country{China}
}

\author{Mingyang Yi}
\email{yimingyang@ruc.edu.cn}
\affiliation{%
  \institution{Renmin University of China}
  \city{Beijing}
  \country{China}
}

\author{Yuyu Luo}
\email{yuyuluo@hkust-gz.edu.cn}
\affiliation{%
  \institution{HKUST (GZ)}
  \city{Guangzhou}
  \country{China}
}

\author{Guoliang Li}
\email{liguoliang@tsinghua.edu.cn}
\affiliation{%
  \institution{Tsinghua University}
  \city{Beijing}
  \country{China}
}

\author{Bin Wu}
\email{wubeen@gmail.com}
\affiliation{%
  \institution{Alibaba Cloud Computing}
  \city{Hangzhou}
  \country{China}
}

\author{Wenchao Zhou}
\email{zwc231487@alibaba-inc.com}
\affiliation{%
  \institution{Alibaba Cloud Computing}
  \city{Hangzhou}
  \country{China}
}

\renewcommand{\shortauthors}{Yuxin Zhang et al.}
\begin{abstract}
Recent advances in large language models (LLMs) trained with reinforcement learning (RL) have improved \nlsql performance. However, RL-based approaches still struggle with complex queries due to two key limitations: insufficient stepwise execution-aware reasoning grounded in database feedback, and the lack of process-level rewards for guiding reasoning optimization.
To address these issues, we propose \cocte, a divide-and-conquer and execution-aware reasoning framework that progressively composes SQL queries through intermediate view validation and structured Common Table Expressions (CTEs), improving both accuracy and interpretability.
To realize a \cocte reasoning process, we develop \sys, a unified approach with three stages: (1) model initialization, which equips LLMs with structured \cocte reasoning capabilities; (2) process reward design, which delivers fine-grained, execution-aware supervision; and (3) process-supervised RL and inference, which integrates process rewards into training and guides the inference stage by process rewards. 
This paper addresses the core challenges in \sys and makes the following contributions. We introduce a process reward model (PRM) that combines execution-aware trajectory scoring with entropy-based step weighting, providing dense and interpretable supervision across reasoning steps. We integrate PRM into both RL training and inference stages, stabilizing optimization and improving trajectory exploration with process-level signals.
Experiments show that \sys significantly outperforms baselines with comparable model sizes, and exhibits strong cross-domain generalization.

\end{abstract}



\received{October 2025}
\received[revised]{January 2026}
\received[accepted]{February 2026}

\begin{document}

\setcopyright{cc}
\setcctype{by}
\acmJournal{PACMMOD}
\acmYear{2026} \acmVolume{4} \acmNumber{3 (SIGMOD)} \acmArticle{228}
\acmMonth{6} \acmDOI{10.1145/3802105}



\maketitle

\input{secs/01introduction}

\input{secs/03problem_formulation}
\input{secs/04system_overview}

\input{secs/05reasoning_enhanced}
\input{secs/06process_reward_model}
\input{secs/07training_inference_alignment}
\input{secs/08experiments}
\input{secs/02related_work}
\input{secs/09conclusions}
\begin{acks}
This work was partially supported by the National Natural Science Foundation
of China (Grant Nos.\ 62436010, 62441230, 62506365, and 62402409) and the Scientific Research
Innovation Capability Support Project for Young Faculty (Grant No.\
SRICSPYF-ZY2025001). 
Guoliang Li was supported by National Key R\&D Program of China (2023YFB4503600), NSF of China (62525202, 62232009), Shenzhen Project (CJGJZD20230724093403007), China Railway Science Research Institute Group Co., Ltd, Zhongguancun Lab.
This work was done while Yuxin Zhang was an intern at Alibaba Cloud Computing, and was supported by Alibaba Research Intern Program.
\end{acks}

\newpage
\bibliographystyle{ACM-Reference-Format}
\bibliography{citations/neurips_2025}


\end{document}

%% file: _commands.tex
\definecolor{dkgreen}{rgb}{0,0.6,0}
\definecolor{gray}{rgb}{0.5,0.5,0.5}
\definecolor{mauve}{rgb}{0.58,0,0.82}
\definecolor{blue}{rgb}{0,0,1}

\definecolor{cadmiumgreen}{rgb}{0.0, 0.42, 0.24}
\definecolor{dropred}{rgb}{0.75, 0.22, 0.17}

\newcommand{\stitle}[1]{\noindent{\bf #1}}
\newcommand{\etitle}[1]{\noindent{\underline{\em #1}}}

\newcommand{\ie}{{\em i.e.,}\xspace}

\newcommand{\term}[1]{{\tt #1}}

\definecolor{shadecolor}{RGB}{240,240,240}

%% file: _notations.tex
\newcommand{\nlsql}{Text-to-SQL\xspace}
\newcommand{\sys}{\textsc{Reward-SQL}\xspace}

\newcommand{\ctes}{\textsc{CTE}s\xspace}
\newcommand{\cocte}{\textsc{CoCTE}\xspace}
\newcommand{\coctes}{\textsc{CoCTE}s\xspace}

\newcommand{\revised}[1]{#1}

\definecolor{review1color}{RGB}{0,102,204}    
\definecolor{review2color}{RGB}{204,0,0}      
\definecolor{review3color}{RGB}{255,128,0}    

\newcommand{\reviewone}[1]{#1}
\newcommand{\reviewtwo}[1]{#1}
\newcommand{\reviewthree}[1]{#1}

%% file: secs/01introduction.tex
\section{Introduction}
\label{sec:introduction}

Text-to-SQL translates natural language (NL) questions into executable SQL queries, allowing non-technical users to interact with relational databases~\cite{gu2024zeroNL2SQL,nl2sqlbugs,zhu2025elliesqlcostefficienttexttosqlcomplexityaware,li2025graphneuralnetworksdatabases, gu2023few,zhang2025taiji,fan2024cost}. 
Despite recent advances in large language models (LLMs), generating complex SQL queries based on LLMs (see a recent survey~\cite{liu2024survey}) still remains a significant challenge, particularly for queries involving multi-table joins and nested structures over complex databases~\cite{lei2024spider,zhang2021tadoc}.

\stitle{RL-based Approaches and Their Limitations.}
Recent work has explored using reinforcement learning (RL) as a post-training strategy to enhance LLMs’ reasoning and SQL generation capabilities~\cite{yao2025arctic,pourreza2025reasoning-sql,lyu2025sql-o1, li2026reasoning, zhang2025deepanalyze}.
The core idea is to enhance LLMs' reasoning for \nlsql by decomposing complex query generation into \emph{intermediate reasoning steps} and leveraging \emph{feedback-driven optimization} to better align model outputs with accurate SQL queries. To achieve this, these methods typically employ policy optimization algorithms, such as Grouped Reinforcement Policy Optimization (GRPO)
in Reasoning-SQL~\cite{pourreza2025reasoning-sql} and Direct Preference Optimization (DPO) in ExCoT variants~\cite{zhai2025excot}, to guide models through structured and step-by-step reasoning.
Despite these advances, current RL-based Text-to-SQL approaches face two key limitations:

\etitle{(1) Limited Stepwise Execution-Aware Reasoning.}
Most RL-based \nlsql approaches~\cite{pourreza2025reasoning-sql,lyu2025sql-o1,yao2025arctic,liu2025dpo-sql} enhance model reasoning by introducing intermediate reasoning steps before generating the final SQL query. 
However, these steps are not \emph{execution-aware}: the model does not interact with the database during generation and only observes inconsistencies after the final SQL is executed. The lack of \emph{stepwise execution signals} limits the model's performance on complex SQL generation. 

\etitle{(2) Lack of Process-Supervised Rewards.}
Existing approaches \citep{pourreza2025reasoning-sql,yao2025arctic,liu2025dpo-sql} primarily rely on outcome-supervised rewards, where feedback is provided after the final SQL query is generated. 
While such rewards encourage execution correctness, the lack of fine-grained, \emph{process-level supervision} hinders effective adjustment of the reasoning trajectory and leads to error accumulation in complex queries.

\begin{figure}[t]
    \centering
    \includegraphics[width=0.65\textwidth]{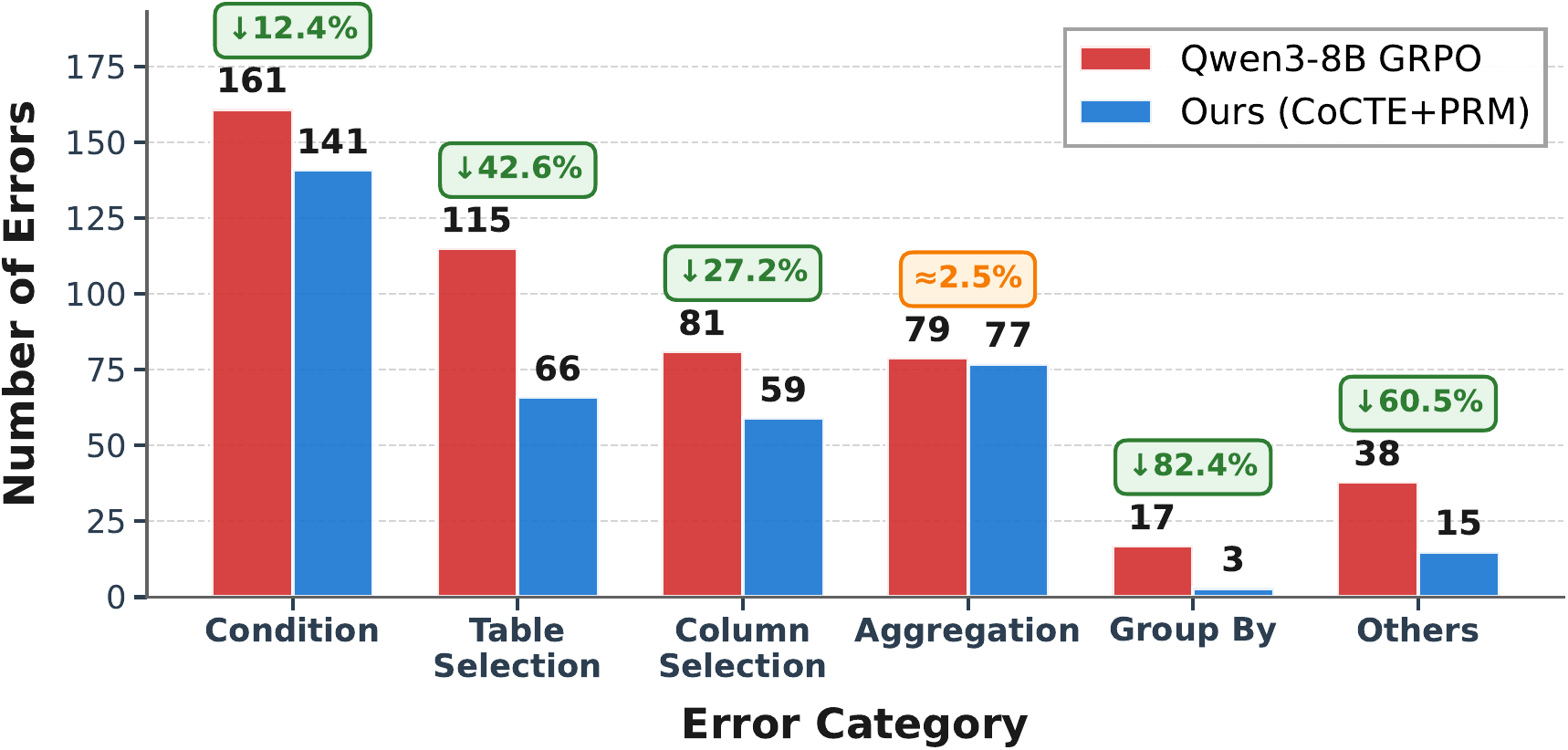}
    \vspace{-0.1in}
    \caption{\reviewone{Error distribution comparison between Qwen3-8B GRPO baseline (red) and our approach (blue).
    }}
    \label{fig:qwen3_error_distribution}
    \vspace{-1em}
\end{figure}

\reviewone{
\stitle{Empirical Evidence from Existing RL Approaches.}
To empirically validate these limitations, we conduct a preliminary study by training a Qwen3-8B model using GRPO with outcome-only rewards on the BIRD training set. We found this baseline achieved only 63.0\% execution accuracy on BIRD Dev. set.
}
\reviewone{
Figure~\ref{fig:qwen3_error_distribution} shows the error distribution of the GRPO baseline. The most frequent errors involve incorrect filter conditions (161 cases, 32.8\%), schema linking errors including table selection (115 cases, 23.4\%) and column selection (81 cases, 16.5\%). These failures stem from a common root cause: the model generates the entire SQL query in one pass without observing intermediate database states. Without stepwise execution feedback, the model cannot verify whether its table joins are correct, or whether filtering conditions align with actual data distributions. This limited ability to ground reasoning in intermediate execution results leads to cascading errors that only become apparent after the final query fails.
}

\begin{figure*}[t]
    \centering
    \subfigure[Traditional Reasoning Framework: introducing intermediate reasoning steps before generating the final SQL query.
    ]{
        \includegraphics[width=0.9\textwidth]{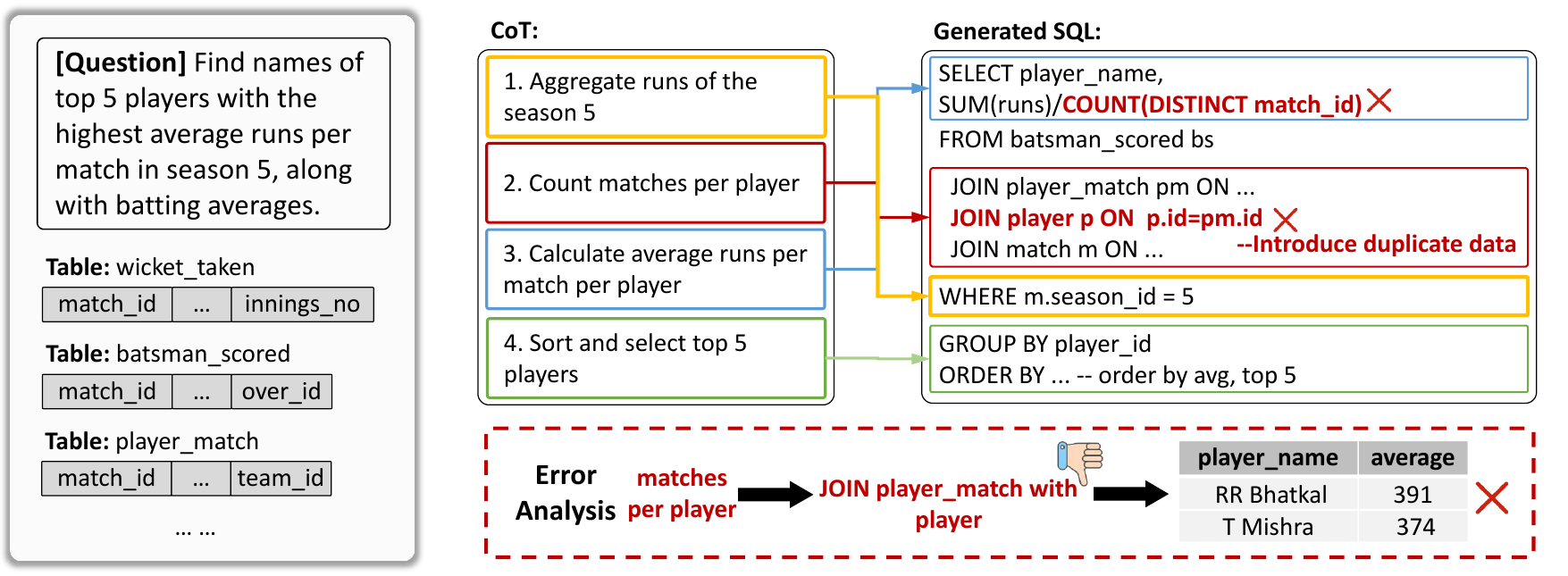} 
        \label{subfig:intro_example_1}
    }
    \vspace{10pt} 
    \subfigure[Our Proposed \cocte: breaking a structurally complex SQL query into a sequence of short and executable
subqueries, \ie \ctes.]{
        \includegraphics[width=0.9\textwidth]{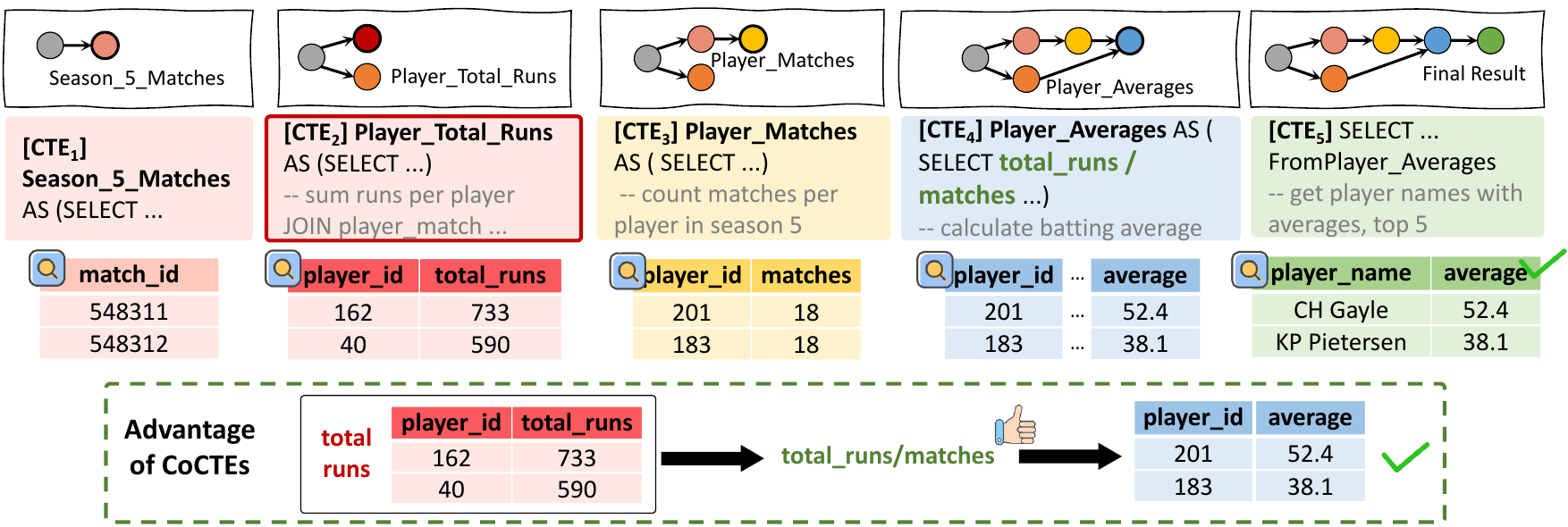} 
        \label{subfig:intro_example_2}
    }
    \vspace{-2em}
    \caption{\reviewone{A real example from the Spider 2.0 benchmark (id:local026-4)~\cite{lei2024spider} comparing traditional Chain-of-Thought (CoT) reasoning by Qwen3-8B GRPO baseline with our proposed 
    framework \cocte.}}
    \label{fig:intro_example}
    \vspace{-1em}
\end{figure*}

\reviewone{
\begin{example}
Figure~\ref{fig:intro_example}(a) illustrates a real failure case from our Qwen3-8B GRPO baseline. The NL question asks for the top 5 players with the highest average runs per match in season 5. The model generates logically sound reasoning steps: (1) aggregate runs by player, (2) count matches per player, (3) calculate average, (4) sort and select top 5.
However, the generated SQL contains a critical error: it unnecessarily joins the \texttt{player} table, causing \textbf{duplication of run counts} when aggregating. Specifically, the formula \texttt{SUM(runs) / COUNT(DISTINCT match\_id)} fails because the erroneous JOIN inflates the numerator while the denominator cannot correct this over-counting. 
This mistake stems from the model's inability to observe intermediate results during generation. Without executing an intermediate step to verify the join result, the model cannot detect that its JOIN strategy produces duplicates. By the time the final SQL executes and returns incorrect results, it is too late to identify which step failed—despite the reasoning being logically sound. $\square$
\end{example}
}

\stitle{\cocte: A Divide-and-Conquer and Execution-Aware Reasoning Framework}.
We propose \cocte, a divide-and-conquer and execution-aware reasoning framework for complex SQL generation. 
The key idea is inspired by how experienced database engineers write SQL queries: instead of producing a single and complex SQL query at once, they build SQL progressively, i.e., \textbf{creating intermediate views to validate intermediate results and structuring logic with Common Table Expressions (CTEs)~\cite{airbyte_cte_guide,adv_sql_concepts_dataforge}}. Each CTE serves as a building block that can be validated and reused, enabling the final query to be built by systematically composing verified components rather than generating a complex SQL query at once. 

Following this idea, our \cocte decomposes a complex query into a sequence of executable \ctes. Specifically, each step is executed as it is generated, providing \emph{stepwise execution-aware feedback} that grounds reasoning in database. This feedback enables \emph{process-level supervision}, allowing the model to detect and correct errors during generation, thus significantly improving the accuracy and interpretability of complex SQL generation.

\begin{example}
    Figure~\ref{fig:intro_example}(b) illustrates the step-by-step reasoning of \cocte for example in Figure ~\ref{fig:intro_example}(a).
    \cocte divides the complex SQL generation process into four executable steps: (1) $\term{CTE}_1$ filters season 5 matches from base tables. (2) $\term{CTE}_2$ computes total runs per player from base tables. (3) $\term{CTE}_3$ counts matches per player based on $\term{CTE}_1$. (4) $\term{CTE}_4$ calculates averages per player from $\term{CTE}_2$ and $\term{CTE}_3$. Finally, the model generates an SQL query to obtain player names with the highest average runs from the results of $\term{CTE}_4$. 
    
    Through stepwise execution-aware feedback, \cocte observes that each player participates in multiple matches and computes the average as $\term{total\_runs}/\term{matches}$, a subtle but crucial step missed by existing methods. By decomposing a structured complex SQL query into a sequence of 
    subqueries, \cocte simplifies generation while improving accuracy and interpretability. $\square$
%
\end{example}

\stitle{Our Proposed \sys Approach.}
To effectively realize the proposed \cocte reasoning framework, we present \sys, a unified approach that integrates stepwise execution-aware reasoning with process-supervised learning. Specifically, \sys is designed with three stages for complex SQL generation.

\etitle{Stage 1 - Model Initialization}. 
    Unlike standard instruction-following, off-the-shelf LLMs lack the ability to follow \cocte-style reasoning. To fix this, \sys synthesizes \cocte-formatted training data and fine-tunes the LLM to learn structured reasoning trajectories, providing a desired initialization for downstream RL process. 
    
\etitle{Stage 2 - Process Reward Design}. We design reliable process-level rewards that deliver fine-grained, execution-aware supervision at each reasoning step. Instead of providing feedback after the SQL is completely generated, these rewards are tied to intermediate execution signals, enabling the model to align reasoning with database.

\etitle{Stage 3 - Process-Supervised RL and Inference.}
    We integrate the designed process rewards into the RL stage, enabling execution-aware supervision throughout training while maintaining alignment with final outcomes. During inference, these rewards guide both efficient selection and structured exploration of reasoning trajectories, enhancing the model's ability to generate accurate queries.



\stitle{Challenges and Solutions.}
Effectively realizing the proposed \sys solution raises two key technical challenges.

The first challenge is designing a process rewards tailored to \cocte reasoning. As discussed above, process rewards aim to overcome the limitations of sparse outcome-based signals in existing RL-based Text-to-SQL approaches~\cite{yao2025arctic,pourreza2025reasoning-sql,lyu2025sql-o1}.
To achieve this goal, we define a process reward model (PRM) that evaluates intermediate trajectories generated within \cocte.
This design introduces two challenges: (1) estimating rewards for each intermediate trajectory based on execution-aware feedback from the corresponding CTE, and (2) weighting rewards across trajectories according to their relative contributions to the final query.
%
To address the challenges, we design a composite process reward consisting of:
(1) a trajectory scoring model that estimates the correctness of intermediate trajectories, enabling step-level reward estimation; and
(2) inverse entropy weighting that emphasizes informative steps and allocates rewards across trajectories accordingly.

The second challenge is integrating the process reward into RL training, which requires a careful balance between the dense reward signal and training stability. An improper combination results in divergent output or reward hacking \citep{shao2024deepseekmath}, where the model over-optimizes high process scores instead of producing the correct final answers.
To address this, during RL, we introduce a unified objective that properly integrates the process and outcome reward signals, leading to a correct and stable training process. Moreover, in inference stage, we use process rewards to enable Best-of-N sampling to select high-quality trajectories.

\stitle{\reviewone{Effectiveness of Our Approach.}}
\reviewone{
As in Figure~\ref{fig:qwen3_error_distribution}, our \sys approach substantially reduces errors across all categories through two complementary mechanisms. \emph{Stepwise execution-aware reasoning} validates schema choices at each CTE, reducing table selection errors by 42.6\% and column selection errors by 27.2\%. \emph{Process-supervised rewards} provide fine-grained feedback during generation, proving particularly effective for complex operations like GROUP BY (82.4\% reduction). 
Together, these mechanisms transform sparse outcome feedback into dense step-level supervision.
}

\stitle{Contributions.}
We summarize our contributions as follows.

(1) We introduce \cocte, a \emph{divide-and-conquer, execution-aware reasoning framework} that decomposes complex queries into a sequence of executable \ctes (Section~\ref{sec:nlcte}).

(2) We propose \sys to effectively realize \cocte by addressing the key technical challenges of process reward design and process-supervised RL and inference (Sections~\ref{sec:prm} and \ref{sec:rl_inference}).

(3) We conduct extensive experiments on standard \nlsql benchmarks. Results show that \sys achieves superior performance on BIRD, reaching 70.3\% execution accuracy with an 8B model and outperforming baselines with comparable parameter sizes. Moreover, \sys demonstrates strong cross-domain generalization, maintaining high performance on five out-of-distribution benchmarks without retraining (Section~\ref{sec:experiments}). \footnote{Code is available at \url{https://github.com/ruc-datalab/RewardSQL}} 


%% file: secs/03problem_formulation.tex
\section{Preliminaries}
\label{sec:preliminaries}

\subsection{Text-to-SQL}
\label{subsec:task}

The Text-to-SQL task maps a natural language (NL) question $q \in \mathcal{Q}$ and a database schema $\mathcal{S}$ to an executable SQL query $y \in \mathcal{Y}$. We denote the schema as $\mathcal{S} = (\mathcal{T}, \mathcal{C}, \mathcal{R})$, where $\mathcal{T}$, $\mathcal{C}$, and $\mathcal{R}$ are the sets of tables, columns, and inter-table relations, respectively. Given $(q, \mathcal{S})$, a \nlsql model generates $y$, which is then executed on a database instance $\mathcal{D}$.
For example, as shown in Figure~\ref{fig:intro_example}, given the question ``\emph{Find names of the top 5 players with the highest average runs per match in season 5, along with batting averages}'', a \nlsql model reasons over multiple tables and constructs a complex SQL query involving filtering, joining, and aggregation, which can be challenging due to multi-table joins and nested structures.


\subsection{Common Table Expressions (CTEs)}
\label{subsec: cte preliminary}

A \textbf{Common Table Expression (CTE)} defines a temporary, named result set within a single SQL query, enabling modular and structured construction of complex queries. Conceptually, a CTE functions similarly to a lightweight, inline \emph{view} as it is defined once and can be referenced multiple times within the same query, without requiring persistent schema changes.
%
A minimal form is:
\begin{verbatim}
WITH cte_name AS ( SELECT ... )
SELECT ... FROM cte_name;
\end{verbatim}
By encapsulating intermediate results as reusable logical views, CTEs allow a complex query to be decomposed into smaller, verifiable subqueries. This modular structure aligns naturally with the stepwise reasoning representation adopted in our \cocte framework, making it easier to verify intermediate steps and compose them into the final query.


\subsection{\reviewone{Design of \cocte}}
\label{sec:design of cocte}


\reviewone{
Recently, Chain-of-Thought (CoT)~\citep{wei2022cot} has been applied to Text-to-SQL to encourage a step by step reasoning. However, existing CoT-based approaches~\citep{li2025alpha-sql,yao2025arctic,pourreza2025reasoning-sql, fan2024autoprep} decompose reasoning solely in the natural language, without grounding intermediate steps in database feedback. As in Figure~\ref{fig:intro_example}, this lack of execution awareness prevents the model from verifying whether each step reflects the actual database state, causing reasoning errors to accumulate and ultimately leading to incorrect SQL generation.
Overcoming this limitation requires a reasoning representation that bridges linguistic abstraction and relational composition in SQL, enabling semantic subgoals to be immediately realized as a verifiable SQL fragments.}

%

\stitle{\reviewone{Formalization of \cocte.}}
\reviewone{To this end, we introduce Chain of Common Table Expressions (\textbf{\cocte}). As mentioned previously, a CTE defines a named, temporary result set within a query, which serves as a modular and interpretable computation unit. Because each CTE can be independently executed and verified, it naturally aligns with how human engineers construct queries, by decomposing problems into smaller, verifiable components.}

\reviewone{
Formally, a \cocte trajectory interleaves natural language reasoning with executable SQL fragments:
$$\mathcal{T} = \{(r_1, q_1), (r_2, q_2), \ldots, (r_k, q_k), (r_{f}, q_{f})\}$$
where $r_i$ is the natural language rationale for step $i$ and $q_i$ is the corresponding CTE. The final query $q_f$ composes all intermediate results to produce the answer.
}

\reviewone{
In Figure~\ref{fig:intro_example}, \cocte decomposes the query generation process into modular, stepwise reasoning steps:
$\text{CTE}_1$ retrieves season 5 matches,
$\text{CTE}_2$ computes total runs per player,
$\text{CTE}_3$ counts matches per player, and
$\text{CTE}_4$ calculates averages before the final selection.
By executing and composing these intermediate subqueries, the model maintains a correct logical flow, and mitigates structural errors common in single-pass 
generation.
%
}

\subsection{Rewards in Reinforcement Learning}
\label{subsec:reward}

In reinforcement learning (RL), the objective is to optimize a policy through reward signals that measure task performance.
The following two primary forms of reward are commonly used.

\textbf{Outcome Reward (OR).} 
The model receives a single reward after completing the entire task, reflecting only the final success or failure. Formally, $R_{\text{out}}$ provides binary or scalar feedback based on the final outcome. While this reward aligns well with the end objective, its sparse and delayed nature makes credit assignment difficult, as it offers no signal about which intermediate decisions contributed to success or error.

\textbf{Process Reward (PR).}
Unlike outcome rewards, process rewards provide step-level feedback that evaluates intermediate actions or reasoning quality. Each step $a_i$ in a trajectory receives a reward $r_i = f_\phi(a_i, a_{<i}) \in [0,1]$, enabling the model to assign credit to specific parts of the reasoning process. This dense supervision improves learning efficiency and stabilizes optimization, but also requires reliable step evaluation to avoid introducing reward bias.

%% file: secs/04system_overview.tex
\section{Overview of \textsc{Reward-SQL}}
\label{sec:overview}

\begin{figure}[t]
    \centering
    \includegraphics[width=0.65\textwidth]{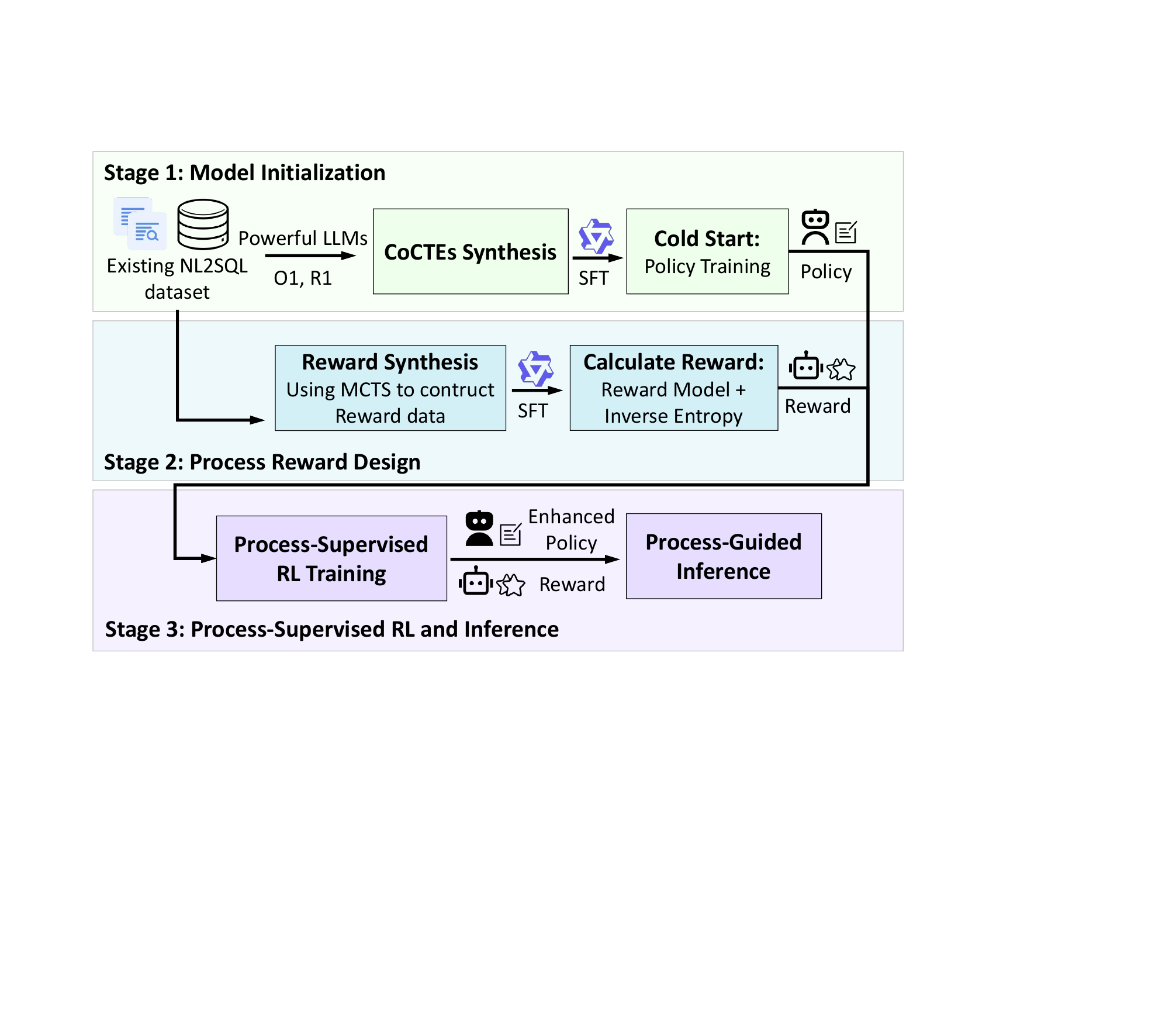}
    \caption{Overview of the \sys framework. It consists of three stages: (1) Model Initialization, which enables the model to generate structured CoCTE-style reasoning; (2) Process Reward Design, which provides stepwise rewards for intermediate trajectories; and (3) Process-Supervised RL and Inference, which integrates process rewards into both training and inference.
    }
    \label{fig:overview}
    \vspace{-1em}
\end{figure}

Figure~\ref{fig:overview} presents our proposed \sys approach, which unifies SQL-level decomposition with process rewards to enhance both training and inference for Text-to-SQL.
Given an NL question $q$ and a database schema $\mathcal{S}$, \sys generates an executable SQL query by constructing a structured \cocte trajectory $\mathcal{T}$. To this end, \sys consists of the following three stages.


\textbf{Stage 1: Model Initialization.}
We first introduce Chain-of-CTEs (namely \cocte), a structured reasoning format that incrementally constructs complex SQL queries through a sequence of intermediate \ctes.
Unlike conventional CoT, which remains purely linguistic, \cocte grounds each reasoning step in an executable subquery, enabling stepwise verification and modular query construction.
Since standard instruction-following LLMs do not inherently produce such structured trajectories, we construct a dedicated training corpus through a semi-automatic pipeline: a strong generator (\textsc{DeepSeek-V3.1}) converts existing NL2SQL queries from \textsc{BIRD}~\cite{li2023bird} and \textsc{Spider}~\cite{yu2018spider} into \cocte trajectories, guided by a small set of manually annotated seed examples.
The policy model $\pi_\theta$ is then initialized via supervised fine-tuning on this corpus, equipping it with the ability to generate coherent, stepwise \cocte trajectories rather than a single SQL query.
This structured generation provides a natural foundation for applying step-level supervision in the subsequent RL training stages (see Section~\ref{sec:nlcte} for details).


\textbf{Stage 2: Process Reward Design.}
Building on \cocte, we construct a \emph{process reward} $R_{\text{proc}}$ that delivers process supervision by aggregating intermediate correctness with uncertainty-aware weighting, which consists of two complementary components:
\begin{itemize}
    \item \textbf{Trajectory Score Model ($R_\phi$)} is trained using labels generated via Monte Carlo Tree Search (MCTS) to estimate the correctness of each intermediate trajectory $\mathcal{T}_{\leq i}$. Specifically, for each NL question, MCTS explores multiple full \cocte trajectories and propagates the terminal execution outcomes back through the search tree to compute Monte Carlo estimates of step-level score for each intermediate node.
    \item \textbf{Inverse Entropy Weight ($IH$)} measures each intermediate trajectory’s contribution to reducing reasoning uncertainty. 
    Trajectories with lower entropy (i.e., higher certainty) receive larger weights, emphasizing informative steps over redundant yet valid ones.
\end{itemize}
By combining $R_\phi$ and $IH$, $R_{\text{proc}}$ converts sparse outcome-only feedback into process-supervised signals, enabling fine-grained supervision throughout the reasoning process (see Section~\ref{sec:prm}).


%

\textbf{Stage 3: Process-Supervised RL Training and Inference.}
We integrate the process reward into both the RL training and inference.
During training, $R_{\text{proc}}$ is combined with the outcome reward $R_{\text{out}}$ under the GRPO framework, providing stepwise supervision for policy optimization.
During inference, $R_{\text{proc}}$ acts as a dynamic guidance signal: in Best-of-$N$ sampling, it selects the trajectory with the highest process reward among $N$ candidates.
The above integration of process rewards allows the model to benefit from process-level supervision throughout both training and inference.
More details are provided in Section~\ref{sec:rl_inference}.

%% file: secs/05reasoning_enhanced.tex
\section{Cold Start for \cocte Policy Model}
\label{subsec:cold_start}\label{sec:nlcte}
As discussed in Section \ref{sec:design of cocte}, \cocte provides a better reasoning paradigm. However, LLMs will not output this pattern as expected. Oppositely, the existing Text-to-SQL model outputs either a single SQL statement or a SQL generated after CoT that consists of natural language. Thus, to adapt LLM to the \cocte format, we apply a \textit{cold-start supervised training process} to the policy model (a LLM) $\pi_{\theta}$ on structured \cocte trajectories.

\textbf{Semi-Automatic Corpus Construction.}
We construct the training corpus through a pipeline as below. \reviewone{Given the natural-language question $q \in \mathcal{Q}$, database schema $\mathcal{S}$, and gold SQL query $y^\star \in \mathcal{Y}$ from the training set as inputs,} \reviewone{w}e first create some handcrafted \cocte as examples to prompt a strong generator (\textsc{DeepSeek-V3.1})~\citep{guo2025deepseekr1} to transform SQL queries from the \textsc{Bird}~\citep{li2023bird} and \textsc{Spider}~\citep{yu2018spider} training sets into \cocte trajectories.
To improve coverage and diversity, each input question is sampled three times at temperature $1.0$, producing multiple candidate trajectories that explore different reasoning paths. Each generated output is then parsed into an abstract syntax tree (AST) and filtered to ensure structural consistency—removing cases with malformed CTE blocks, undefined references, or naming conflicts. 
The resulting dataset provides syntactically and semantically valid \cocte trajectories.

Table~\ref{tab:coldstart-stats} presents statistics of the constructed corpus. 
From 17,462 initial SQL queries, the pipeline produces 51,996 valid \cocte trajectories with a 90.7\% acceptance rate after filtering.
The average trajectory contains 3.3 CTE steps, with rationales averaging 28.1 tokens and CTEs averaging 11.6 tokens, demonstrating the granular decomposition achieved by the \cocte format.

\begin{table}[h]
\centering
\small
\vspace{-0.1in}
\caption{Statistics of the \cocte corpus constructed for cold-start training. 
The dataset combines \textsc{Bird} and \textsc{Spider} queries transformed via a semi-automatic pipeline.}
\vspace{-1mm}
\label{tab:coldstart-stats}
\begin{tabular}{l l r}
\toprule
\textbf{Category} & \textbf{Metric} & \textbf{Value} \\
\midrule
\multirow{4}{*}{Data Scale}
  & \# Manually annotated seeds & 5 \\
  & \# Initial SQL queries & 17{,}462 \\
  & \# Valid \coctes (post-filtering) & 51{,}996 \\
  & Acceptance rate & 90.7\% \\
\midrule
\multirow{4}{*}{Structure Statistics}
  & Avg. CTE steps & 3.3 \\
  & Max. CTE steps & 9 \\
  & Avg. rationale length & 28.1 tokens \\
  & Avg. CTE length & 11.6 tokens \\
\bottomrule
\end{tabular}
\vspace{-1mm}
\end{table}

\textbf{Supervised Fine-Tuning}
We fine-tune the policy model $\pi_\theta$ on this constructed corpus to imitate the reasoning pattern of \cocte by minimizing a next-token prediction loss~\citep{jaech2024openai}:
\[
\mathcal{L}_{\text{SFT}}(\theta)
= 
-\sum_{t} \log \pi_\theta((r_t^\star, q_{t}^{\star}) \mid (r_{<t}^\star, q_{<t}^{\star}), q, \mathcal{S}),
\]
where each $(r_t^\star, q_{t}^{\star})$ denotes a paired intermediate step of NL rationale and executable CTE.

By doing so, the policy model generates coherent, executable \cocte trajectories instead of isolated SQL or textual explanations.
The model learns to alternate between semantic planning (rationales) and structural realization (CTEs), producing outputs that are both interpretable and executable. More importantly, such structured \cocte provides a natural basis for step-level supervision: each intermediate CTE can be independently executed and evaluated, enabling the process reward design described in the next Section. 

%% file: secs/06process_reward_model.tex
\section{Process Reward Design}
\label{sec:prm}
\subsection{Motivation and Formulation}

The existing RL approaches for Text-to-SQL mostly rely on the \textit{Outcome Reward} (OR), a binary variable of execution correctness:
\[
R_{\text{out}}(y) =
\begin{cases}
1, & \mathrm{Exe}(y,\mathcal{D}) = \mathrm{Exe}(y^\star,\mathcal{D}),\\
0, & \text{otherwise.}
\end{cases}
\]
Although conceptually straightforward, this reward is extremely sparse and non-decomposable.  
Even if most intermediate CTEs in a \cocte trajectory are correct, a single error in the final clause nullifies the entire signal, providing no clue as to which reasoning step failed.  
This ``all-or-nothing'' feedback amplifies the credit assignment problem and leads to high-variance policy gradients, often scaling super-linearly with the reasoning horizon~\citep{wu2018variance,preiss2019variance}.

To overcome these limitations, we redefine the reward formulation from a \emph{terminal} outcome to a \emph{process-oriented} metric. 
Concretely, we introduce the \textbf{Process Reward}, a dense reward signal that distributes credit across each executable step of a \cocte trajectory.  
For a reasoning trajectory $\mathcal{T}=\{(r_i,q_i)\}_{i=1}^k$, we define the process reward over each intermediate reasoning trajectory $\mathcal{T}_{\leq d} = \{q_{1},\cdots , q_{d}\}$ ($d\leq k$) as:
\begin{equation}\label{eq:process_reward}
R_{\text{proc}}(\mathcal{T}_{\leq d}) = 
\sum_{i=1}^{d} \left(\frac{\sum_{j\leq i}IH(\mathcal{T}_{\leq j}) \cdot R_\phi(\mathcal{T}_{\leq j})}{\sum_{j\leq i}IH(\mathcal{T}_{\leq j})}\right).    
\end{equation}

Intuitively, this formulation is an \textbf{uncertainty-reweighted sum} of intermediate trajectory correctness, where:
\begin{itemize}
    \item $R_\phi(\mathcal{T}_{\leq j})$: \textbf{Trajectory Score Model} that estimates the correctness of intermediate trajectory $\mathcal{T}_{\leq j}$
    \item $IH(\mathcal{T}_{\leq j})$: \textbf{Inverse Entropy Weight} that quantifies the contribution of trajectory $\mathcal{T}_{\leq j}$ to reducing  uncertainty
\end{itemize}

This design ensures that our process reward captures both \emph{semantic correctness} (via $R_\phi$) and \emph{reasoning efficiency} (via $IH$).
Steps that are both correct and informative receive higher credit, while redundant or uninformative steps are automatically down-weighted.
Next, we describe how each component is constructed.

\subsection{Trajectory Score Model}
\label{subsec:trajectory score model}
To estimate the quality of CTE steps, we train a Trajectory Score Model $R_\phi$ that maps the reasoning trajectory to a confidence score:
\[
s_i = R_\phi(\mathcal{T}_{\leq i} \mid  q, \mathcal{S}) \in (0,1).
\]
A higher $s_i$ indicates stronger confidence that the current CTE $q_i$ is locally valid and globally consistent with the target query. 
The core challenge lies in obtaining accurate step-level supervision without human annotation.  
Inspired by the use of Monte Carlo estimation in mathematical reasoning~\citep{luo2024OmegaPRM}, we adopt a Monte Carlo Tree Search (MCTS) strategy to generate step-level labels automatically.

Concretely, for each question in the \textsc{Bird} training set, we sample $n$ trajectories from the cold-started policy model $\pi_{\mathrm{ref}}$ and perform MCTS to explore the space of \cocte completions. 

Each node in the search tree corresponds to an intermediate trajectory $\mathcal{T}_{\le i}$, and the leaf nodes represent complete SQL candidates whose execution yields binary rewards $R_{\text{out}}\in\{0,1\}$.  
\reviewone{To ensure sufficient diversity, we use $n=8$ rollouts with temperature $T=1.0$, yielding an average of 38 distinct root-to-leaf trajectories per question, effectively preventing collapse to repeated paths.}
By backing up these terminal outcomes through the tree, we compute a Monte Carlo estimate of each intermediate step's expected utility:
\[
\sigma_i = \mathbb{E}_{\pi_{\mathrm{ref}}}[R_{\text{out}} \mid \mathcal{T}_{\le i}]
     = \!\!\!\sum_{\mathcal{T}_{\le i} \subseteq \tau}\!\! 
        p(\tau\mid\mathcal{T}_{\le i})\, R_{\text{out}}(\tau).
\]
Intuitively, $\sigma_i\in[0,1]$ can be viewed as the success rate from intermediate trajectory $\mathcal{T}_{\leq i}$ to the end. 
To mitigate noise in $\sigma_{i}$, we apply syntax-level filters to remove malformed CTEs and resolve name conflicts, ensuring structural validity.

In our \cocte structure, each CTE step is paired with its execution output using \texttt{SELECT * FROM} queries, exposing both syntactic structure and runtime semantics to the model.  
The model input thus interleaves each CTE fragment, its execution result, and a \texttt{<step>} marker token indicating the prediction checkpoint.  
Given these step-labeled samples, the Trajectory Score Model $R_{\phi}(\mathcal{T}_{\leq i}\mid q, \mathcal{S})$ is optimized via binary cross-entropy:
\[
\mathcal{L}_{\text{PRM}}
= - \sum_{i=1}^{k} 
  \big[\sigma_i \log s_i + (1-\sigma_i)\log(1-s_i)\big].
\]

By doing so, the trajectory score model provides correctness estimation on reasoning trajectories.  \reviewthree{We implement the Trajectory Score Model $R_\phi$ using Qwen2.5-7B-Coder-Instruct as the base architecture, training it independently without parameter sharing with the policy model.}
However, correctness is insufficient to capture a step's \emph{usefulness}. 
For example, a redundant step can be correct in syntactic sense, while it may fail to advance the overall reasoning (e.g., redundant joins or irrelevant filters).

\subsection{Inverse Entropy Weight}
\label{subsec: inverse entropy weight}

During the construction of the training set, we observe that the Trajectory Score Model indeed sometimes rewards \textit{spurious but valid} steps—CTEs that are syntactically correct and executable yet fail to advance the overall reasoning.  
To distinguish genuinely contributive steps from merely valid ones, we incorporate an additional process-level signal derived from the entropy of policy model.

At each generation step, the entropy of the policy model over next tokens $\pi_\theta(\cdot \mid \mathcal{T}_{\leq i})$ is 
\[H(a\mid \mathcal{T}_{\leq i}) = -\!\!\sum_a \pi_\theta(a \mid \mathcal{T}_{\leq i})
      \log \pi_\theta(a \mid \mathcal{T}_{\leq i}),\]
\reviewone{where $a$ denotes a single token from the vocabulary}, which measures the model's uncertainty over the next token: a lower $H(a\mid \mathcal{T}_{\leq i})$ corresponds to less uncertainty.  
Empirically, we find that entropy tends to decrease monotonically along successful reasoning steps as in Figure \ref{fig:reward_example}.  

Based on this observation, we propose to assign higher weights to informative (low entropy) intermediate steps. 
We define the entropy of step $q_{i + 1} = (a_{i + 1}^{1},\cdots, a_{i + 1}^{|q_{i + 1}|})$ as:
\[H(q_{i + 1}\mid \mathcal{T}_{\leq i}) = \frac{1}{|q_{i + 1}|}\sum_{j=1}^{|q_{i + 1}|}H\left(a_{i + 1}^{j}\mid \mathcal{T}_{\leq i}, a_{i + 1}^{<j}\right),\]
where $|q_{i + 1}|$ is the sequence length. Then, the entropy of intermediate trajectory $\mathcal{T}_{\leq i} = \{q_{1},\cdots, q_{i}\}$ is defined as $H(\mathcal{T}_{\leq i}) =  \sum_{j=1}^{i}H(q_{j}\mid \mathcal{T}_{< j})$. 

We then normalize the trajectory entropy as $\tilde{H}(\mathcal{T}_{\leq i}) = \frac{H(\mathcal{T}_{\leq i})}{\max_{i}H(\mathcal{T}_{\leq i})}$, and define the \emph{Inverse Entropy}:
\begin{equation}
IH(\mathcal{T}_{\leq i}) = 1 - \tilde{H}(\mathcal{T}_{\leq i}).
\end{equation}
A larger inverse entropy $IH(\mathcal{T}_{\leq i})$ indicates a more certain intermediate trajectory, meaning the prior steps have effectively constrained the reasoning space and provided strong guidance for subsequent generation.
This inverse entropy serves as the weight in Equation~\ref{eq:process_reward}, ensuring that informative steps contribute more to the overall process reward.

\subsection{Summary and Example}

By combining the Trajectory Score Model $R_{\phi}$ with the Inverse Entropy weight $IH$, the process reward design achieves two complementary goals:
(1) verifiable correctness through the trajectory scoring model $R_{\phi}$, and
(2) reward allocation that reflects the exploratory contribution of each intermediate step via the $IH$ weights.



\begin{figure}[t]
    \centering
    \includegraphics[width=0.7\textwidth]{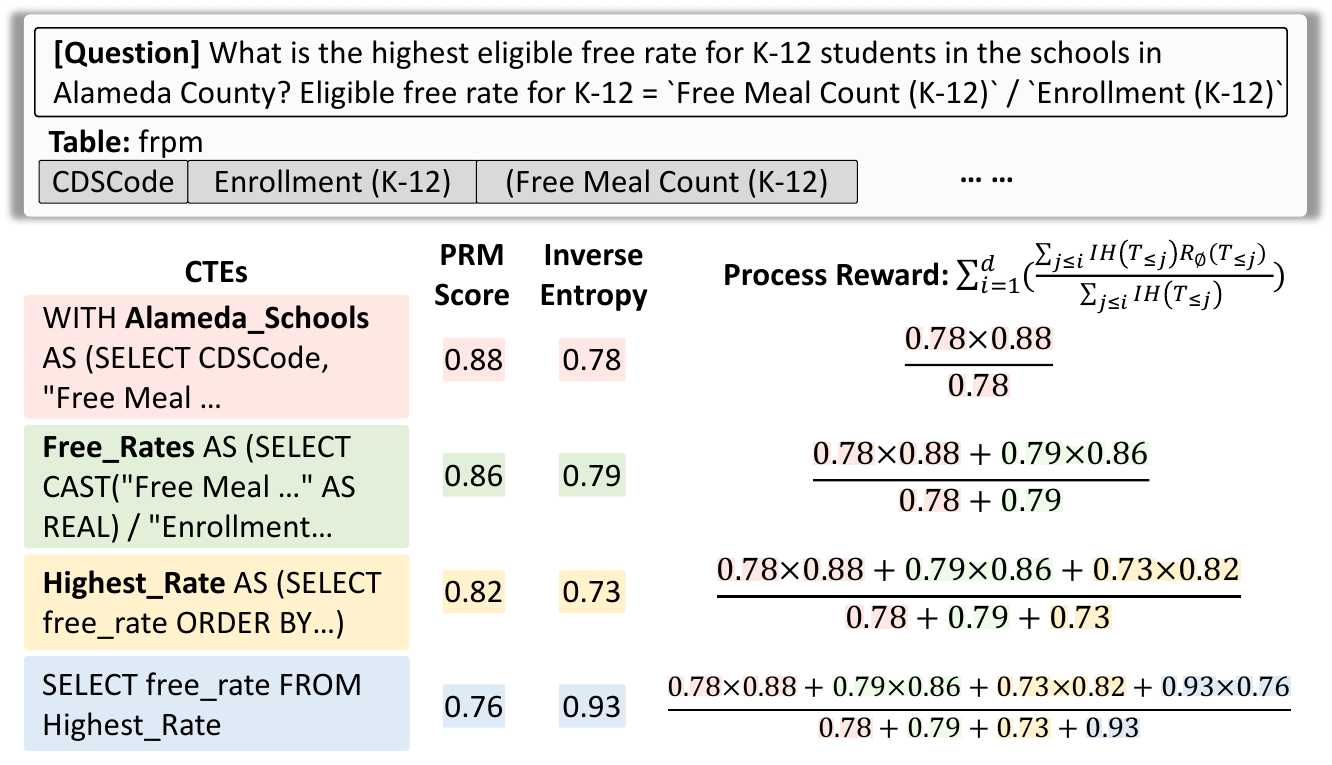}
    \caption{The computation of process reward, which is an
inverse entropy weighted sum of trajectory scores.}
    \label{fig:reward_example}
\end{figure}

\begin{example}
Figure~\ref{fig:reward_example} illustrates the process reward computation for the Alameda County query. 
The trajectory consists of four CTE steps with their corresponding Trajectory Scores and entropy values and following Equation~\ref{eq:process_reward}, we can compute the process reward at each step.
Notably, the first CTE (\texttt{Alameda\_Schools}) has both high correctness (0.88) and low inverse entropy (0.78), indicating it is both valid and informative, thus receiving strong positive signal.
In contrast, \texttt{Highest\_Rate} has slightly lower correctness and higher entropy, resulting in reduced contribution to the overall reward.
The final SELECT step exhibits very high inverse entropy (0.93), reflecting high certainty after all intermediate reasoning steps.
\end{example}

\subsection{\reviewtwo{Decomposition Quality Analysis}}
\label{subsec:decomp_quality}
\reviewtwo{
While \cocte trajectories provide verifiable intermediate results, they may also introduce structural redundancy or non-optimal decompositions. To understand this phenomenon and validate our inverse-entropy weighting mechanism, we manually analyzed 500 randomly sampled trajectories from the cold-start corpus.
}

\reviewtwo{
\stitle{Redundancy Statistics.}
We categorize trajectories into two groups: those containing redundant CTEs (e.g., duplicate joins, unnecessary filtering steps) and those with minimal redundancy. As shown in Table~\ref{tab:redundancy_analysis}, approximately 37\% (184/500) of trajectories exhibit some form of structural redundancy. These include cases where intermediate CTEs perform semantically equivalent operations or fail to progressively narrow the result set.
}

\begin{table}[t]
\centering
\caption{\reviewtwo{Entropy statistics for 500 sampled trajectories with and without redundancy. Redundant trajectories exhibit higher entropy, which our inverse-entropy weighting mechanism down-weights during training.}}
\label{tab:redundancy_analysis}
\small
\scalebox{0.92}{\begin{tabular}{lcc}
\toprule
\textbf{Metric} & \textbf{With Redundancy} & \textbf{Without Redundancy} \\
\midrule
Sample Count & 184 & 316 \\
Avg. Entropy & 0.3142 & 0.2966 \\
Avg. Inverse Entropy & 0.47 & 0.50 \\
\bottomrule
\end{tabular}}
\end{table}

\reviewtwo{
\stitle{Entropy as Quality Indicator.} 
Crucially, we observe that redundant trajectories exhibit higher average entropy (0.3142 vs. 0.2966) and correspondingly lower inverse entropy (0.47 vs. 0.50). This validates our design intuition: when the policy model generates redundant steps, it exhibits greater uncertainty in token-level predictions, reflected in elevated entropy values. By incorporating inverse-entropy weighting in Equation~\ref{eq:process_reward}, our PRM automatically assigns lower credit to such uncertain steps, mitigating their negative impact during RL training.
}

%% file: secs/07training_inference_alignment.tex
\section{Process-Supervised RL and Inference}
\label{sec:rl_inference}


\subsection{Process-Supervised RL Training}
As demonstrated in recent work~\citep{li2025alpha-sql,li2025omnisql,lyu2025sql-o1,ma2025sql-r1,pourreza2025reasoning-sql}, reinforcement learning post-training significantly improves Text-to-SQL capabilities. 
However, as discussed in Section~\ref{sec:introduction}, existing RL approaches rely on sparse Outcome Reward (OR) signals that provide only binary feedback on final execution correctness.
This sparsity leads to unstable training dynamics and poor credit assignment, particularly for multi-step \cocte 
trajectories.

To address this limitation, we integrate our process reward (Equation~\ref{eq:process_reward}) into the RL optimization objective, providing dense, step-level supervision throughout the reasoning process.

\textbf{Unified Process Reward Objective.}
\label{subsec:process_reward_objective}
Formally, the RL optimization objective is maximizing the expected reward:
\begin{equation}\label{eq:process_reward_objective}
J(\theta) = 
\mathbb{E}_{(\mathcal{T}, y) \sim \pi_\theta(\cdot\mid q, \mathcal{S})}
\left[
R_{\text{proc}}(\mathcal{T}) + R_{\mathrm{out}}(y) - \lambda D_{KL}(\pi_{\theta}\parallel \pi_{\mathrm{ref}})
\right],
\end{equation}
Here $D_{KL}(\pi_{\theta}\parallel \pi_{\mathrm{ref}})$ penalizes the policy model away from the initialized pre-trained model $\pi_{\mathrm{ref}}$, and $\lambda > 0$ is a hyperparameter.   
This formulation transforms reinforcement learning from sparse outcome-level supervision into dense, interpretable, step-level guidance.  
Each reasoning step in an \cocte trajectory contributes its own verifiable credit, allowing gradients to flow through intermediate reasoning processes rather than being determined solely by final execution success.

\textbf{Integration into GRPO.}
We conduct the RL process under the GRPO framework \citep{shao2024deepseekmath}, which has been proven to be an effective and stable RL training framework. 

Given query $(q, \mathcal{S})$, the GRPO  samples $G$ trajectories $\{(\mathcal{T}_{g}, y_{g})\}_{g=1}^{G}$ from the current policy model $\pi_\theta(\mathcal, y\mid q, \mathcal{S})$. Then  
for every token $a_{g, i}^{j}$ in step $q_{g, i}$, we define the reward 
\[A_{g, i} = R_{out}(y_{g}) + \sum_{k=i}^{d} \left(\frac{\sum_{j\leq k}IH(\mathcal{T}_{g, \leq j})R_\phi(\mathcal{T}_{g, \leq j})}{\sum_{j\leq k}IH(\mathcal{T}_{g, \leq j})}\right),\]
and standardize it across the group as $\hat{A}_i = 
\frac{A_{i} - \mu_R}{\sigma_R}$, 
where $\mu_R$ and $\sigma_R$ are the mean and standard deviation of all $A_{i}$ in groups.  
The GRPO maximizes the following loss:
\begin{equation}
\begin{aligned}
\mathcal{L}_{\text{GRPO}} & =\\ 
& \frac{1}{G}\sum_{g=1}^{G}\frac{1}{\sum |a_{g, i}^{j}| + |y_{g}|}\sum_{q_{g, i}\in \mathcal{T}_{g}}\sum_{j}
\min
\left(
\rho_j \hat{A}_j,\,
\text{clip}(\rho_j,1\pm\epsilon)\hat{A}_j
\right) \\
& - \lambda D_{KL}(\pi_{\theta}\parallel \pi_{\mathrm{ref}}),
\label{eq:grpo_objective}
\end{aligned}
\end{equation}
where $\rho_{j} = \pi_{\theta}(a_{g, i}^{j}\mid q, \mathcal{S}, \mathcal{T}_{<g}, a_{g, i}^{<j}) / \pi_{\mathrm{ref}}(a_{g, i}^{j}\mid q, \mathcal{S}, \mathcal{T}_{<g}, a_{g, i}^{<j})$, the KL divergence is estimated as in \citep{shao2024deepseekmath}, and $\sum |a_{g, i}^{j}|$ represents the number of tokens in trajectory $\mathcal{T}_{g}$. 
During training, the PRM parameters $\phi$ remain frozen, while only policy $\pi_\theta$ is updated.
This converts process-level feedback into a differentiable supervision signal, enabling stable credit assignment with reduced variance.

\begin{figure}[t!]
    \centering
    \includegraphics[width=0.65\textwidth]{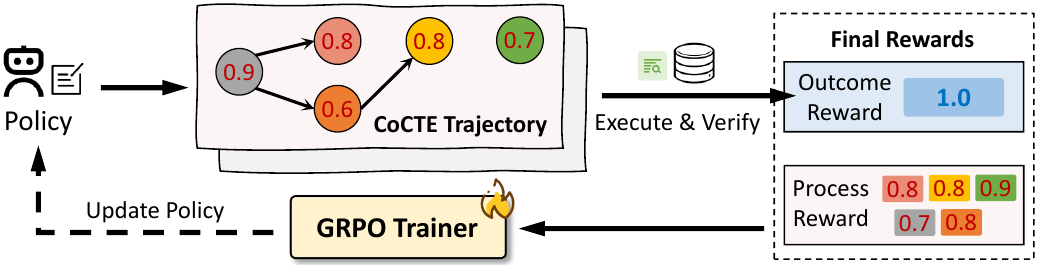}
    \caption{The GRPO with process reward signals integrated.}
    \label{fig:process_supervised_rl_training}
\end{figure}

\subsection{Process-guided Inference}
\textbf{Inference-Time Process Guidance.}
Beyond training, the process reward also serves as a principled mechanism for inference-time guidance.
During inference, reasoning trajectories are generated auto-regressively, and the process reward provides step-level quality estimates that can guide trajectory selection and exploration.
Unlike existing Text-to-SQL systems that rely on heuristic methods such as majority voting or self-consistency~\citep{dong2023c3}, our approach leverages the trained process reward to evaluate and rank candidate reasoning paths based on their semantic correctness and reasoning efficiency.

We integrate the process reward into a widely used inference strategies: Best-of-$N$ sampling.

\textbf{Best-of-$N$ Decoding.}
The policy model $\pi_\theta$ generates $N$ \cocte trajectories for a given query.
Each candidate trajectory $\mathcal{T}$ is assigned a process reward $R_{\text{proc}}(\mathcal{T})$ by \eqref{eq:process_reward}.
Then, the trajectory with the highest process reward is selected as the final output:
\[
\mathcal{T}^* = \arg\max_{\mathcal{T} \in \{\mathcal{T}_1, \ldots, \mathcal{T}_N\}} R_{\text{proc}}(\mathcal{T}).
\]
This selection strategy replaces the heuristic voting mechanisms with a principled, learned evaluation metric that accounts for both correctness and reasoning quality.


In summary, the process reward serves as a dynamic inference-time controller that enables principled trajectory selection.
By replacing static heuristics with learned, step-level evaluations, our approach completes a closed loop where the model learns from process supervision during training and reasons with process evaluation during inference.

%% file: secs/08experiments.tex
\section{Experiments}
\label{sec:experiments}

As summarized in Figure~\ref{fig:overview}, our \sys framework consists of three stages: 
(1)~\emph{Cold start}, which guides the policy model to generate reasoning in the \cocte format; 
(2)~\emph{Process reward-assisted RL post-training}, which improves the generation probability on trajectories with higher process rewards; and 
(3)~\emph{Process reward-assisted inference}, which leverages process rewards to select high-quality trajectories during inference.
In this section, we conduct a comprehensive empirical evaluation of the proposed \sys.

\subsection{Experimental Setup}
\label{subsec:setup}

\stitle{Evaluation Datasets.}
We evaluate \sys~on a suite of benchmarks covering both cross-domain and domain-specific SQL reasoning.
For standard evaluation, we adopt two representative large-scale cross-domain datasets:
\textsc{BIRD}~\cite{li2023bird} and \textsc{Spider}~\cite{yu2018spider}.
\textsc{BIRD} features industrial-scale databases with rich numerical and textual attributes (9{,}428 train, 1{,}534 dev, 1{,}789 test), where database schemas are strictly disjoint across splits.
\textsc{Spider} follows the same cross-domain split protocol (7{,}000 train, 1{,}034 dev, 2{,}147 test) and remains the de-facto benchmark for complex multi-table SQL generation.
Unless otherwise specified, all reported results are measured on the filtered BIRD development and Spider test sets.

\reviewone{
\stitle{Generalization Benchmarks.}
To assess model generalization beyond the training distribution, we evaluate on five out-of-distribution (OOD) benchmarks spanning two levels of domain shift:
}
\reviewone{
\emph{Robustness-level generalization} is tested on three Spider-based variants that preserve schema familiarity while introducing linguistic variations:
\textbf{Spider-DK}~\cite{gan2021spider-dk} (535 samples) tests inference of implicit domain knowledge;
\textbf{Spider-Syn}~\cite{gan2021spider-syn} (1{,}034 samples) replaces schema mentions with synonyms to examine lexical robustness; and
\textbf{Spider-Realistic}~\cite{deng2020spider-realistic} (508 samples) uses real-user questions with practical ambiguities.
}
\reviewone{
\emph{Cross-domain generalization} is evaluated on two domain-specific benchmarks with entirely different schemas and specialized terminology:
\textbf{ScienceBenchmark}~\cite{zhang2023sciencebenchmark} (299 samples) covers research policy, astrophysics, and cancer studies with scientific databases, while
\textbf{EHRSQL}~\cite{lee2022ehrsql} (1{,}008 samples) focuses on clinical databases requiring medical domain expertise.
None of these five benchmarks are used during training, representing strict zero-shot evaluation.
}

\stitle{Model and Training Configuration.}
We adopt \textsc{Qwen3-8B} as the base model for all experiments.
Training proceeds in three consecutive stages:
(1)~\emph{Supervised fine-tuning} (SFT) on the \cocte trajectories introduced in Section~\ref{sec:nlcte}, enabling the model to generate stepwise reasoning and compositional SQL structures;
(2)~\emph{Online reinforcement learning} using the unified reward $R_{\mathrm{proc}}+R_{\mathrm{out}}$ under the GRPO framework (Section~\ref{sec:rl_inference}); and
(3)~\emph{Process-guided inference} using the Process Rewards (\ref{sec:prm}) as the selection criterion during decoding.
The SFT stage employs a learning rate of $1\times10^{-5}$, batch size~64, maximum context length~4{,}096, and 3 epochs of training.
For GRPO, the group size $G$ is set to~8, clipping ratio~0.2, KL coefficient~0.02, and sampling temperature~1.0.
During inference, the model generates $N$ trajectories and ranks them with process scores; $N\!=\!8$ is used by default.

\stitle{Metrics and Decoding Strategies.}
Model quality is evaluated by \emph{Execution Accuracy (EX)}, which compares the outputs of predicted and reference SQL queries executed on the target database.
Two decoding protocols are used:
(1)~greedy decoding with temperature zero, and
(2)~PRM-guided Vote@$N$, where multiple candidates are generated and scored by their process reward.

\stitle{Implementation and Environment.}
All models are implemented in PyTorch~2.6.0 with DeepSpeed ZeRO-3 optimization. 
Experiments are conducted on a \textsc{Rocky~Linux~8.10} server equipped with four \textsc{NVIDIA~H20-3e} GPUs (143\,GB each), dual \textsc{Intel~Xeon~Platinum~8468V} processors (96~cores in total), and 2\,TB system memory. 
Evaluation is performed on the same hardware to ensure consistent latency measurements.

\subsection{Main Results: Comparison with Text-to-SQL Baselines}
\label{subsec:baselines}

We first compare \sys with existing specialized Text-to-SQL approaches on standard benchmarks.
Table~\ref{tab:main_comparison} presents results on BIRD development set and Spider test set, which are the two most widely used benchmarks for evaluating Text-to-SQL systems.

\stitle{Baselines.}
We compare with three main categories of specialized Text-to-SQL approaches:
\reviewtwo{(1)~\emph{Structure-aware methods} that encode database schema and foreign key relations into graph neural networks, including RAT-SQL~\cite{wang2019rat} and RESDSQL~\cite{li2023resdsql};}
(2)~\emph{Prompting-based methods} that leverage large language models with carefully designed prompts for schema linking and query construction, including DIN-SQL~\cite{Pourreza2023dinsql}, DAIL-SQL~\cite{Gao2024dailsql}, MAC-SQL~\cite{Wang2025macsql}, Chase-SQL~\cite{Pourreza2024chasesql}, and Alpha-SQL~\cite{li2025alpha-sql};
\reviewtwo{(3)~\emph{Post-training methods} includes supervised fine-tuning approaches such as CodeS~\cite{li2024codes}, DTS-SQL~\cite{Pourreza24dtssq;}, SQL-o1~\cite{lyu2025sql-o1} and Omni-SQL~\cite{li2025omnisql}, as well as RL-based methods such as SQL-R1~\cite{ma2025sql}, Reasoning-SQL~\cite{pourreza2025reasoning-sql}, and Arctic-SQL~\cite{yao2025arctic}.}

\begin{table}[t]
\centering
\caption{\reviewtwo{Comparison with Text-to-SQL methods on benchmarks. The methods evaluated under Vote@$N$ (self-consist. or selection agent or MCTS) are marked with $^\ddagger$. For open-source methods, we reproduce the results ourselves; for closed-source methods, we report results from original papers. Best results are \textbf{bolded}.}}

\label{tab:main_comparison}
\scalebox{0.872}{\begin{tabular}{lcc}
\toprule
\textbf{Model / Method} & \textbf{BIRD} & \textbf{Spider} \\
& \textbf{(dev)} & \textbf{(test)} \\
\midrule
\multicolumn{3}{l}{\textit{\reviewtwo{Structure-aware Methods}}} \\
\reviewtwo{RAT-SQL + GraPPa} & - & 73.3 \\
\reviewtwo{RESDSQL + T5-3B} & - & 79.9 \\
\midrule
\multicolumn{3}{l}{\textit{Prompting-based Methods}} \\
DIN-SQL + GPT-4 & 50.7 & 85.3 \\
DAIL-SQL + GPT-4 & 54.8 & \textbf{86.2} \\
MAC-SQL + GPT-4 & 59.4 & - \\
Alpha-SQL + Qwen2.5-7B$^\ddagger$ & 66.8 & 84.0 \\
GPT-4o & 61.9 & - \\
\reviewtwo{openPangu 7B} & 41.9 & - \\
\midrule
\multicolumn{3}{l}{\textit{\reviewtwo{Supervised Fine-tuning Methods}}} \\
DTS-SQL + DeepSeek-7B & 56.0 & - \\
CodeS + StarCoder-15B & 58.5 & 79.4 \\
SQL-o1 + Qwen2.5-7B$^\ddagger$ & 66.7 & 85.1 \\
\reviewtwo{Omni-SQL + Qwen2.5-7B$^\ddagger$} & 63.8 & 79.4 \\
\midrule
\multicolumn{3}{l}{\textit{\reviewtwo{RL-based Methods}}} \\
Reasoning-SQL + Qwen2.5-7B & 64.0 & 78.7 \\
\quad + Vote@$N^\ddagger$ & 68.1 & - \\
\reviewtwo{SQL-R1 + Qwen2.5-7B$^\ddagger$} & 65.1 & 81.4 \\
\reviewtwo{Arctic-SQL + Qwen2.5-7B$^\ddagger$} & 66.8 & 82.1 \\
\midrule
\multicolumn{3}{l}{\textit{Our Approach (Qwen3-8B)}} \\
\sys + Greedy & 65.9 & 81.2 \\
\sys + PRM@8 (Best-of-N)$^\dagger$ & 68.5 & 83.1 \\
\sys + PRM@32 (Best-of-N)$^\dagger$ & \textbf{70.3} & 85.4 \\
\bottomrule
\end{tabular}}
\end{table}

\stitle{Results and Analysis.}
The results in Table~\ref{tab:main_comparison} demonstrate several key advantages of our approach:

\reviewtwo{
\textbf{(1) Improved Performance over Structure-aware Models.}
Our \sys improves the results obtained under traditional structure-aware methods like RAT-SQL and RESDSQL, validating the effectiveness of LLM and RL-post training.} 

\reviewtwo{
\textbf{(2) Improved Performance over other LLM-based Methods.} Firstly, for all post-training methods with comparable model sizes (7-8B parameters), \sys achieves the best performances on BIRD and Spider. These improvements are consistent across both greedy decoding and test-time scaling. Besides, even compared to strong (100$\times$ larger) closed-source LLMs (e.g., GPT-4 based methods), our \sys has comparable or improved results. These results together show that the RL-post training combined with our CoCTEs reasoning structures and PRM can make a small LLM have strong performance on Text-to-SQL.}

\reviewone{
\subsection{Generalization Study}
\label{subsec:generalization}
}

\reviewone{
We evaluate generalization on five OOD benchmarks, explicitly categorizing them into two levels:
\begin{itemize}
\item \textbf{Robustness-level OOD}: Spider-DK, Spider-Syn, and Spider-Realistic evaluate robustness to linguistic variations (domain knowledge, synonym substitution, and realistic ambiguity) while retaining schema similarity to the training data.
\item \textbf{Cross-domain-level OOD}: ScienceBenchmark and EHR-SQL evaluate transfer to entirely unseen domains with different schemas and specialized terminology. 
\end{itemize}
}

\reviewone{
\stitle{Setup.}
We compare against three categories of models to ensure fair comparisons:
(1)~\emph{Specialized 7-8B models} that are post-trained on Text-to-SQL data, including reproducible RL-based methods (SQL-R1, Arctic-SQL) and SFT methods (Omni-SQL);
(2)~\emph{General-purpose models} of similar size (7-8B): Qwen2.5-Coder-7B and Meta-Llama-3.1-8B;
(3)~\emph{Large-scale models}: Qwen2.5-72B, DeepSeek-V3, GPT-4-Turbo, and GPT-4o.
For fair comparison, Vote@8 results for all open-source models use self-consistency voting.}

\begin{table*}[t]
\centering
\caption{\reviewone{Out-of-distribution generalization across five challenging benchmarks. We distinguish two levels of OOD evaluation: \textbf{robustness-level} and \textbf{cross-domain-level}. Results show execution accuracy (\%) for greedy decoding and sampling-based selection. Best results in each category are in bold.}}

\label{tab:generalization}
\small
\setlength{\tabcolsep}{3pt}
\scalebox{0.76}{
\begin{tabular}{l|@{\hspace{6pt}}cc@{\hspace{6pt}}cc@{\hspace{6pt}}cc@{\hspace{8pt}}cc|cc@{\hspace{6pt}}cc@{\hspace{8pt}}cc|cc}
\toprule
\multirow{3}{*}{\textbf{Model}} & \multicolumn{8}{c|}{\textbf{\reviewone{Robustness-Level OOD}}} & \multicolumn{6}{c|}{\textbf{\reviewone{Cross-Domain-Level OOD}}} & \multicolumn{2}{c}{\multirow{2}{*}{\textbf{Overall}}} \\
\cmidrule(lr){2-9} \cmidrule(lr){10-15}
& \multicolumn{2}{c}{\textbf{Spider-DK}} & \multicolumn{2}{c}{\textbf{Spider-Syn}} & \multicolumn{2}{c}{\textbf{Spider-Real}} & \multicolumn{2}{c|}{\textbf{Avg.}} & \multicolumn{2}{c}{\textbf{ScienceBench}} & \multicolumn{2}{c}{\textbf{EHR-Bench}} & \multicolumn{2}{c|}{\textbf{Avg.}} & \multicolumn{2}{c}{} \\
\cmidrule(lr){2-3} \cmidrule(lr){4-5} \cmidrule(lr){6-7} \cmidrule(lr){8-9} \cmidrule(lr){10-11} \cmidrule(lr){12-13} \cmidrule(lr){14-15} \cmidrule(lr){16-17}
& \multicolumn{1}{c}{\textbf{Gre}} & \multicolumn{1}{c}{\textbf{Maj}} & \multicolumn{1}{c}{\textbf{Gre}} & \multicolumn{1}{c}{\textbf{Maj}} & \multicolumn{1}{c}{\textbf{Gre}} & \multicolumn{1}{c}{\textbf{Maj}} & \multicolumn{1}{c}{\textbf{Gre}} & \multicolumn{1}{c|}{\textbf{Maj}} & \multicolumn{1}{c}{\textbf{Gre}} & \multicolumn{1}{c}{\textbf{Maj}} & \multicolumn{1}{c}{\textbf{Gre}} & \multicolumn{1}{c}{\textbf{Maj}} & \multicolumn{1}{c}{\textbf{Gre}} & \multicolumn{1}{c|}{\textbf{Maj}} & \multicolumn{1}{c}{\textbf{Gre}} & \multicolumn{1}{c}{\textbf{Maj}} \\
\midrule
\multicolumn{17}{l}{\textit{Closed-Source Models}} \\
GPT-4-Turbo & 72.3 & 72.1 & 62.9 & 63.5 & 67.5 & 68.3 & 67.6 & 68.0 & \textbf{59.2} & \textbf{59.5} & 43.1 & 44.8 & \textbf{51.2} & \textbf{52.2} & 61.0 & 61.6 \\
GPT-4o & 72.9 & 73.5 & 59.6 & 62.3 & 66.5 & 66.7 & 66.3 & 67.5 & 55.5 & 56.2 & \textbf{44.9} & \textbf{45.5} & 50.2 & 50.9 & 59.9 & 60.8 \\
\midrule
\multicolumn{17}{l}{\textit{Large Open-Source Models}} \\
Qwen2.5-72B-Instruct & \textbf{76.4} & \textbf{77.6} & 64.1 & 64.3 & 70.1 & 68.5 & 70.2 & 70.1 & 52.8 & 58.2 & 35.0 & 41.2 & 43.9 & 49.7 & 59.7 & 62.0 \\
DeepSeek-V3 (671B MoE) & 72.9 & 73.8 & 64.4 & 65.1 & 67.9 & 66.9 & 68.4 & 68.6 & 56.2 & 57.9 & 43.2 & 43.5 & 49.7 & 50.7 & 60.9 & 61.4 \\
\midrule
\multicolumn{17}{l}{\textit{Similar-Sized Models ($\sim$7-8B)}} \\
Qwen2.5-Coder-7B-Instruct & 67.5 & 73.6 & 63.1 & 66.9 & 66.7 & 70.5 & 65.8 & 70.3 & 45.2 & 51.2 & 24.3 & 36.9 & 34.8 & 44.1 & 53.4 & 59.8 \\
Meta-Llama-3.1-8B-Instruct & 62.6 & 69.9 & 53.1 & 59.3 & 57.5 & 61.0 & 57.7 & 63.4 & 36.8 & 43.1 & 24.6 & 33.7 & 30.7 & 38.4 & 46.9 & 53.4 \\
\midrule
\multicolumn{17}{l}{\textit{\reviewone{Post-Trained Models ($\sim$7-8B)}}} \\
\reviewone{Arctic-SQL + Qwen2.5-7B} & 69.7 & 71.8 & 75.2 & 76.0 & 80.3 & 81.7 & 75.1 & 76.5 & 52.5 & 53.8 & 36.3 & 38.9 & 44.4 & 46.4 & \textbf{62.8} & 64.4 \\
\reviewone{OmniSQL + Qwen2.5-7B} & 64.9 & 68.2 & 70.7 & 73.9 & 75.2 & 80.7 & 70.3 & 74.3 & 50.5 & 51.8 & 34.3 & 39.5 & 42.4 & 45.7 & 59.1 & 62.8 \\
\reviewone{SQL-R1 + Qwen2.5-7B} & 67.9 & 70.5 & 72.6 & 75.4 & 78.3 & 80.5 & 72.9 & 75.5 & 53.8 & 51.8 & 36.0 & 38.8 & 44.9 & 45.3 & 61.7 & 63.4 \\
\midrule
\multicolumn{17}{l}{\textit{Our Model}} \\
\textbf{\sys + Qwen3-8B} & 69.4 & 73.1 & \textbf{75.7} & \textbf{79.0} & \textbf{84.1} & \textbf{83.3} & \textbf{76.4} & \textbf{78.5} & 50.8 & 54.6 & 29.6 & 34.4 & 40.2 & 44.5 & 61.9 & \textbf{64.9} \\
\bottomrule
\end{tabular}
}
\end{table*}

\stitle{\reviewone{Results and Analysis.}}
\reviewone{
Table~\ref{tab:generalization} presents comprehensive results across all five OOD benchmarks.
Our analysis reveals that:
}

\reviewone{\textbf{(1) Robustness-Level OOD: Superior Performance Against RL Baselines.} On Spider-DK, Spider-Syn, and Spider-Realistic, where schemas remain similar but linguistic variations increase, our approach achieves SOTA performance among all 7-8B models. 
Notably, we surpass even 100× larger models including GPT-4-Turbo and DeepSeek-V3. This demonstrates that process-aware training produces superior robustness to linguistic variations compared to both RL approaches and massive general-purpose models.}


\reviewone{\textbf{(2) Cross-Domain-Level OOD: Competitive with RL but Dropped Performance due to Schema Transfer.}
On the ScienceBenchmark and EHR-SQL, where domains and schemas are entirely unseen, the performance drops for all specialized 7-8B models. Though \sys achieves comparable results with other post-trained baselines, all these RL-post trained models significantly underperform large general models on these benchmarks. This is because the post-training process pushes the model to overfit to domain-specific schemas and terminology, which deteriorates the performance of the model on the other domains. Similar phenomena have been observed in \citep{huan2025does}. Handling such OOD generalization is left as a challenge in future work.}


\reviewone{
\textbf{(3) Overall Trade-offs.}
When averaged across all OOD benchmarks, \sys achieves 64.9\%, marginally exceeding Arctic-SQL (64.4\%)—while substantially outperforming SQL-R1 (63.4\%) and OmniSQL (62.8\%). This demonstrates that \sys achieves superior OOD generalization compared to RL methods, while maintaining strong performance across both cross-domain and linguistic variation settings.
}



\textbf{(4) Effectiveness of Test-Time Scaling Across OOD Types.}
PRM-guided selection provides consistent gains in both OOD scenarios, though the magnitude varies. On robustness-level tasks, scaling from greedy to Maj@8 improves performance by +2.1\% on average (76.4\% → 78.5\%), validating that process rewards effectively capture reasoning patterns resilient to linguistic variations. On cross-domain tasks, gains are comparable (+4.3\%, 40.2\% → 44.5\%), suggesting that process-level supervision remains beneficial even when schema knowledge is incomplete, though it cannot fully compensate for domain-specific knowledge gaps.

The consistent improvements across all benchmarks (averaging +9.2\% over the best closed-source model) indicate that combining \cocte structured reasoning with process reward supervision yields robust and generalizable SQL generation rather than dataset-specific overfitting.


\subsection{\revised{Training Stability Analysis}}
\label{subsec:training_alignment}

\revised{
We examine how process-level supervision influences reinforcement learning. Traditional RL post-training for Text-to-SQL relies solely on the \emph{Outcome Reward} $R_{\text{out}}$, a binary signal indicating execution correctness. While simple, this signal is highly sparse: most trajectories receive zero reward, and its gradient variance grows with trajectory length, leading to unstable optimization.
\sys augments the outcome reward with the dense \emph{Process Reward} $R_{\text{proc}}$ defined in Section~\ref{sec:rl_inference}, forming a composite objective $R_{\text{proc}} + R_{\text{out}}$ under the GRPO framework. This design provides credit signals at intermediate steps of a \cocte trajectory rather than only at the final SQL.
%
}

\begin{figure}[H]
\centering
\includegraphics[width=0.7\textwidth]{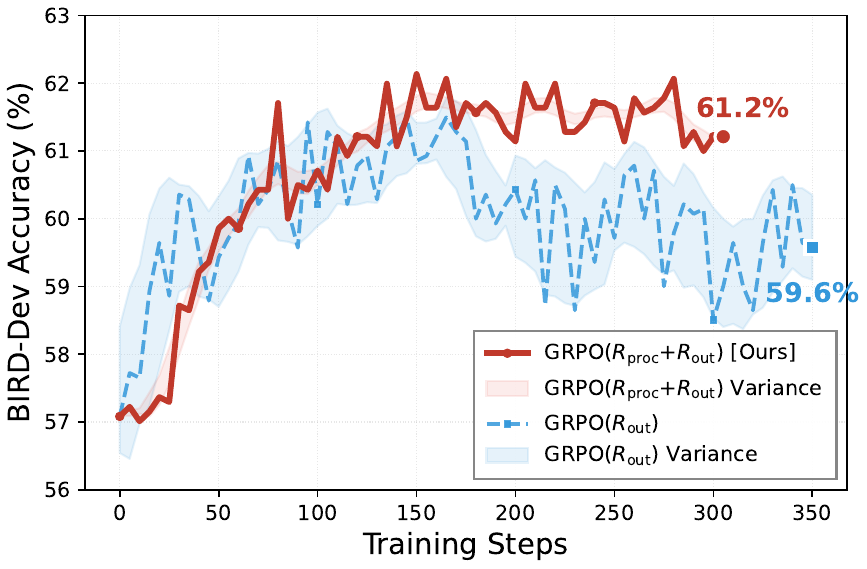}
\caption{\revised{Training dynamics of GRPO with and without process reward.
Process supervision accelerates convergence and reduces variance in dev-set execution accuracy.}}
\label{fig:training_curve}
\end{figure}

\stitle{\revised{Training Dynamics.}}
\revised{
Figure~\ref{fig:training_curve} compares the dev-set execution accuracy during GRPO training using only $R_{\text{out}}$ versus the unified $R_{\text{proc}}{+}R_{\text{out}}$ objective.
Both models start from the same SFT checkpoint and share identical hyperparameters.
In the first 150 training steps, both variants exhibit similar upward trends, with $GRPO(R_{\text{proc}}{+}R_{\text{out}})$ achieving a higher peak performance of 62.1\%. After 150 steps, the training dynamics diverge significantly: the process-guided model maintains near-stable performance with variations not exceeding 1\%, whereas the outcome-only $GRPO(R_{\text{out}})$ variant experiences substantial instability, showing noticeable performance degradation and variance of approximately 1.5\%.
}

\subsection{Ablation Studies}
\label{subsec:ablation}

\stitle{\revised{Quantitative Comparison.}}
\revised{
Table~\ref{tab:grpo_training} summarizes the final performance and training stability.
Under identical configurations, GRPO($R_{\text{proc}}{+}R_{\text{out}}$) achieves 61.8\% execution accuracy on the BIRD development set, outperforming the outcome-only baseline (59.6\%) while exhibiting a substantially lower variance after the initial 50 steps ($\pm$0.58 vs $\pm$0.76).
The number of steps to convergence decreases from 350 to 250, demonstrating that the dense process reward provides a more informative learning signal and facilitates early stabilization of policy gradients.
}

\begin{table}[t!]
\centering
\caption{\revised{Effect of process reward on GRPO training stability. Results show final execution accuracy, mean performance, and standard deviation after initial 50 steps.}}
\label{tab:grpo_training}
\begin{tabular}{lccc}
\toprule
\textbf{Method} & \textbf{Dev EX (\%)} & \textbf{Mean ($\pm$)} & \textbf{Std ($\pm$)} \\
\midrule
GRPO ($R_{\text{out}}$) & 59.6 & 60.2 & 0.76 \\
GRPO ($R_{\text{proc}}{+}R_{\text{out}}$) & \textbf{61.8} & \textbf{61.5} & \textbf{0.58} \\
\bottomrule
\end{tabular}
\end{table}

In this section, we conduct ablation studies on the main components of our \sys. The main results are summarized in Table \ref{tab:ablation}.


\begin{table}[t]
\centering
\caption{\reviewone{Ablation study on key components of the proposed framework. Results compare different reasoning formats, training methods, and selection strategies on BIRD Dev.}}
\label{tab:ablation}
\scalebox{0.9}{
\begin{tabular}{lcc}
\toprule
\textbf{Configuration} & \textbf{Greedy} & \textbf{PRM@8} \\
& \textbf{EX (\%)} & \textbf{EX (\%)} \\
\midrule
Full \sys (SFT + GRPO + PRM@8) & 66.0  & 68.7 \\
\midrule
\textit{Reasoning Format Variants} \\
\quad Direct SQL (no reasoning) & 63.6 & - \\
\quad \reviewone{\cocte (prompting Deepseek-V3.1)} & 60.1 & - \\
\quad NL-CoT (generic reasoning) & 64.2 & - \\
\quad \cocte (ours) & \textbf{66.0} & \textbf{68.7} \\
\midrule
\textit{Training Method Variants} \\
\quad SFT only & 63.2 & 67.4 \\
\quad GRPO($R_{\text{out}}$ only) & 62.6 & 67.0 \\
\quad GRPO($R_{\text{proc}}+R_{\text{out}}$) & \textbf{66.0} & \textbf{68.7} \\
\midrule
\textit{Selection Method Variants} \\
\quad Self-Consistency & - & 66.6 \\
\quad Outcome Reward (Model) & - & 67.1 \\
\quad Process Reward (ours) & - & \textbf{68.7} \\
\bottomrule
\end{tabular}
}
\end{table}

\textbf{(1) Reasoning Format.}
\cocte outperforms both direct SQL generation (66.0\% vs 63.6\% in greedy, +2.4\%) and generic natural language chain-of-thought (66.0\% vs 64.2\%, +1.8\%).
This validates our design of executable intermediate reasoning steps that can be verified against the database, providing more reliable supervision signals than unverifiable natural language reasoning or format-only prompting.
\reviewone{To further validate that gains come from our training/reward design rather than merely using CTE formatting, we compare against 5-shot CoCTE prompting with DeepSeek-V3.1, which achieves only 60.1\%. The 5.9\% gap (66.0\% vs 60.1\%) demonstrates that prompting alone cannot reliably induce proper SQL decomposition—our process-supervised training is essential for learning correct substep generation.}

\textbf{(2) Training Method.}
GRPO training with process rewards improves over SFT baseline by 2.7\% in greedy decoding (66.0\% vs 63.2\%).
More importantly, comparing GRPO($R_{\text{proc}}+R_{\text{out}}$) with GRPO($R_{\text{out}}$) shows that adding process rewards during training improves both greedy (66.0\% vs 62.6\%, +3.4\% absolute improvement) and PRM@8 performance (68.7\% vs 67.0\%, +1.7\% gain).
This demonstrates that process-level supervision during training enhances model's ability to generate high-quality intermediate reasoning steps.

\textbf{(3) Selection Method.}
PRM selection substantially outperforms alternatives, achieving 68.7\% compared to self-consistency (66.6\%, +2.1\% improvement) and ORM selection (67.1\%, +1.6\% gain).
This confirms that process-level evaluation provides more fine-grained discrimination among candidate trajectories than outcome-only or voting-based selection.
Overall, these ablations confirm that all three components—\cocte format, process-aware GRPO training, and PRM-guided selection—contribute to the final performance.

\reviewtwo{
\stitle{Sensitivity to Cold-Start Corpus Source.}
To evaluate whether our policy model overfits to a specific LLM's decomposition style during cold-start training, we conduct experiments using trajectories generated exclusively by DeepSeek, GPT-4, or Claude as the initial training corpus (Section~\ref{subsec:cold_start}). As shown in Table~\ref{tab:llm_sensitivity}, policy models trained on single-source corpora achieve comparable performance
: DeepSeek-only (68.4\% at PRM@8), GPT-only (67.6\%), and Claude-only (68.9\%). 
The small variance (±0.6\%) demonstrates that our trained policy model generalizes across different reasoning styles and does not overfit to any particular LLM's CTE decomposition patterns.
Moreover, training on a mixed corpus combining trajectories from all three LLMs yields further improvements. 
This suggests that exposure to diverse decomposition styles enhances the policy model's ability to generate varied reasoning paths.
}

\begin{table}[t]
\centering
\caption{\reviewtwo{Sensitivity to trajectory source: training on different single-LLM corpora versus mixed-source corpus.}}
\label{tab:llm_sensitivity}
\scalebox{0.9}{
\begin{tabular}{lcc}
\toprule
\textbf{Training Source} & \textbf{Greedy} & \textbf{PRM@8} \\
\midrule
DeepSeek-only & 58.15 & 68.4 \\
GPT-4-only & 58.87 & 67.6 \\
Claude-only & 62.97 & 68.9 \\
\midrule
Mixed (DeepSeek+GPT+Claude) & \textbf{64.86} & \textbf{70.0} \\
\bottomrule
\end{tabular}
}
\end{table}

\begin{table}[t!]
\centering
\caption{\reviewone{Error category analysis on BIRD Dev. Comparison between Qwen3-8B and \sys with different PRM@N.}}
\label{tab:error_analysis}
\resizebox{0.84\textwidth}{!}{%
\begin{tabular}{llccc}
\toprule
\textbf{Category} & \textbf{Error Type} & \textbf{Qwen3-8B} & \textbf{\sys} & \textbf{\sys} \\
& & \textbf{Errors (\%)} & \textbf{PRM@8 (\%)} & \textbf{PRM@32 (\%)} \\
\midrule
\multirow{6}{*}{\textit{Schema Linking}}
& Table Selection & 100 (6.52) & 65 (4.24) & 62 (4.04) \\
& Column Selection & 88 (5.74) & 51 (3.32) & 48 (3.13) \\
& JOIN Keys & 49 (3.19) & 14 (0.91) & 13 (0.85) \\
& Hallucination & 96 (6.26) & 6 (0.39) & 4 (0.26) \\
& Condition Error & 102 (6.65) & 65 (4.24) & 63 (4.11) \\
& NULL Handling & 22 (1.43) & 52 (3.39) & 52 (3.39) \\
\midrule
\multirow{2}{*}{\textit{Value Retrieval}}
& Format Error & 41 (2.67) & 24 (1.56) & 25 (1.63) \\
& Manipulation & 8 (0.52) & 12 (0.78) & 9 (0.59) \\
\midrule
\multirow{3}{*}{\textit{Operation}}
& Math Formula & 43 (2.80) & 46 (3.00) & 34 (2.22) \\
& Aggregation & 42 (2.74) & 37 (2.41) & 36 (2.35) \\
& Complex Operation & 7 (0.46) & 5 (0.33) & 2 (0.13) \\
\midrule
\multirow{3}{*}{\textit{Information}}
& Filter Condition & 29 (1.89) & 47 (3.06) & 53 (3.45) \\
& ORDER BY/LIMIT & 26 (1.69) & 31 (2.02) & 30 (1.96) \\
& Return Format & 12 (0.78) & 13 (0.85) & 11 (0.72) \\
\midrule
\multirow{2}{*}{\textit{Other}} 
& Domain Knowledge & 6 (0.39) & 9 (0.59) & 9 (0.59) \\
& Invalid SQL & 20 (1.30) & 4 (0.26) & 2 (0.13) \\
\midrule
\textbf{Total} & & \textbf{699 (45.57)} & \textbf{483 (31.49)} & \textbf{455 (29.66)} \\
\bottomrule
\end{tabular}
}
\end{table}

\subsection{Computational Cost Analysis}
\label{subsec:cost_analysis}

While test-time scaling improves accuracy, it incurs additional computational costs. We analyze the trade-offs between performance gains and resource requirements, distinguishing between one-time training costs and per-query inference overhead.

\stitle{\reviewtwo{Training-Time vs. Inference-Time Costs.}}
\reviewtwo{
We first clarify that the expensive MCTS search is performed \textit{only once} during offline training-data construction (Section 5.2) and incurs no overhead during inference. Specifically, MCTS sampling processed \textbf{16,248} training instances, consuming approximately \textbf{18 hours} on a single server. The subsequent PRM training utilized \textbf{4$\times$A800 GPUs for 7 hours} (28 GPU-hours total). These one-time offline costs amortize across all future queries.}

\reviewtwo{
For online inference, the computational overhead comprises three stages: (1)~candidate generation via VLLM, (2)~SQL execution to obtain intermediate results, and (3)~PRM scoring for candidate selection. Below we analyze the latency of each component.}

\stitle{Latency Analysis.} Figure~\ref{fig:latency_breakdown} presents a breakdown of the latency components for different values of n in our PRM@n approach. The inference pipeline consists of three main stages:

\begin{figure}[t]
\centering
\includegraphics[width=0.73\textwidth]{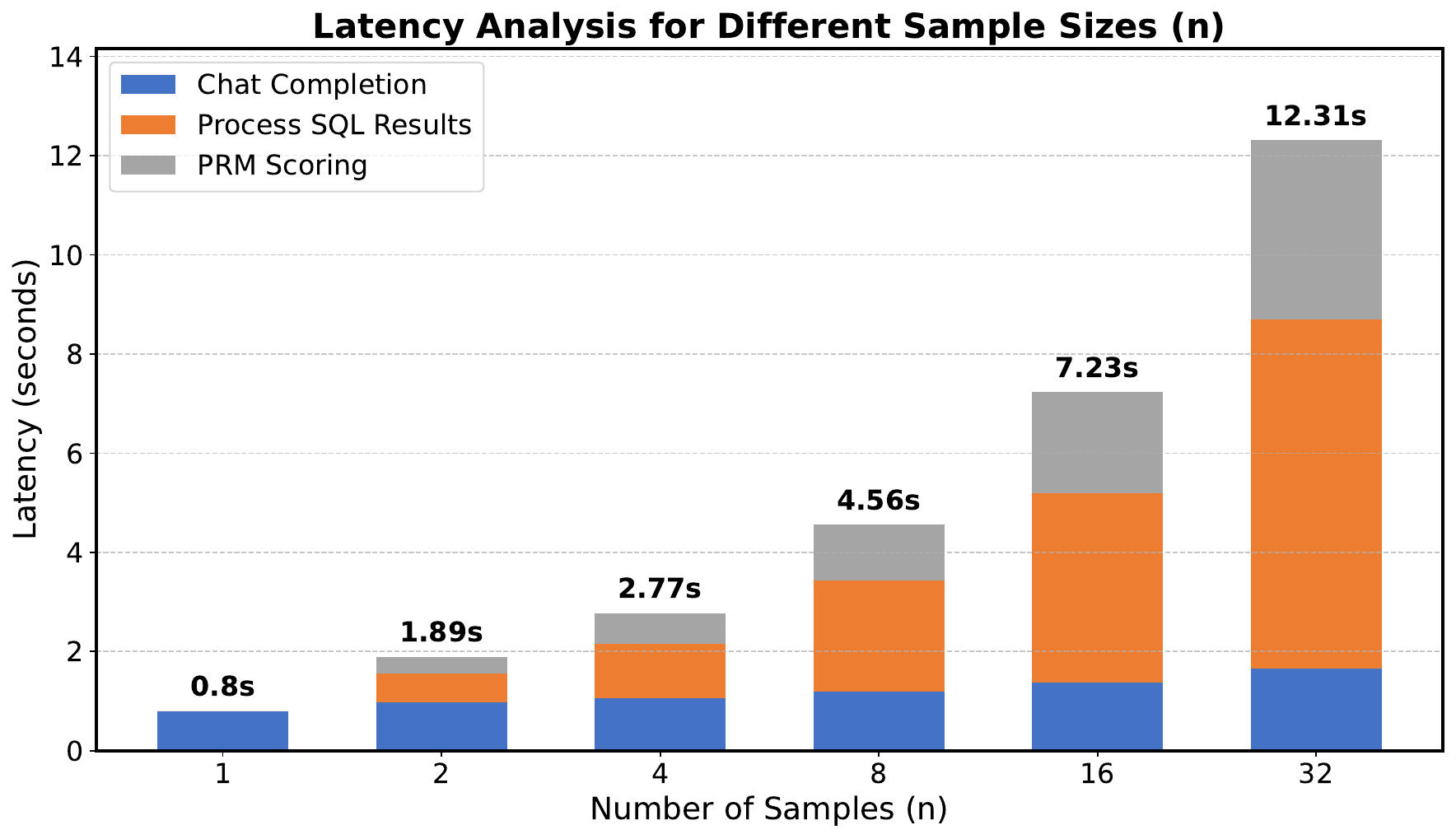}
\caption{Latency analysis for different values of n in PRM@n. The chart shows the time contribution of each component.}
\label{fig:latency_breakdown}
\end{figure}

\begin{itemize}[leftmargin=*]
\item \textbf{Chat Completion:} Leveraging VLLM's optimized batch inference, resulting in sublinear latency growth (from 0.8s at n=1 to 1.65s at n=32)
\item \textbf{SQL Processing:} Collecting intermediate results through execution, scaling linearly with n. \reviewone{In this part, our \sys has similar latencies compared to standard inference methods without our intermediate CoCTEs, due to database cache.}
\reviewone{{\item \textbf{PRM Scoring:} Evaluating candidates with the Process Reward Model, whereas the extra computational costs of our \sys arise from it. Rising linearly from 0.21s-3.61s (depending on n). This part takes up 5\%-20\% cost of total latency.}}
\end{itemize}

The total latency increases from 0.8 seconds at n=1 to 12.31 seconds at n=32, with SQL processing becoming the dominant factor at higher values of n. This analysis reveals significant optimization opportunities, particularly in parallelizing the SQL execution and scoring components.

\begin{figure}[t]
\centering
\includegraphics[width=0.7\textwidth]{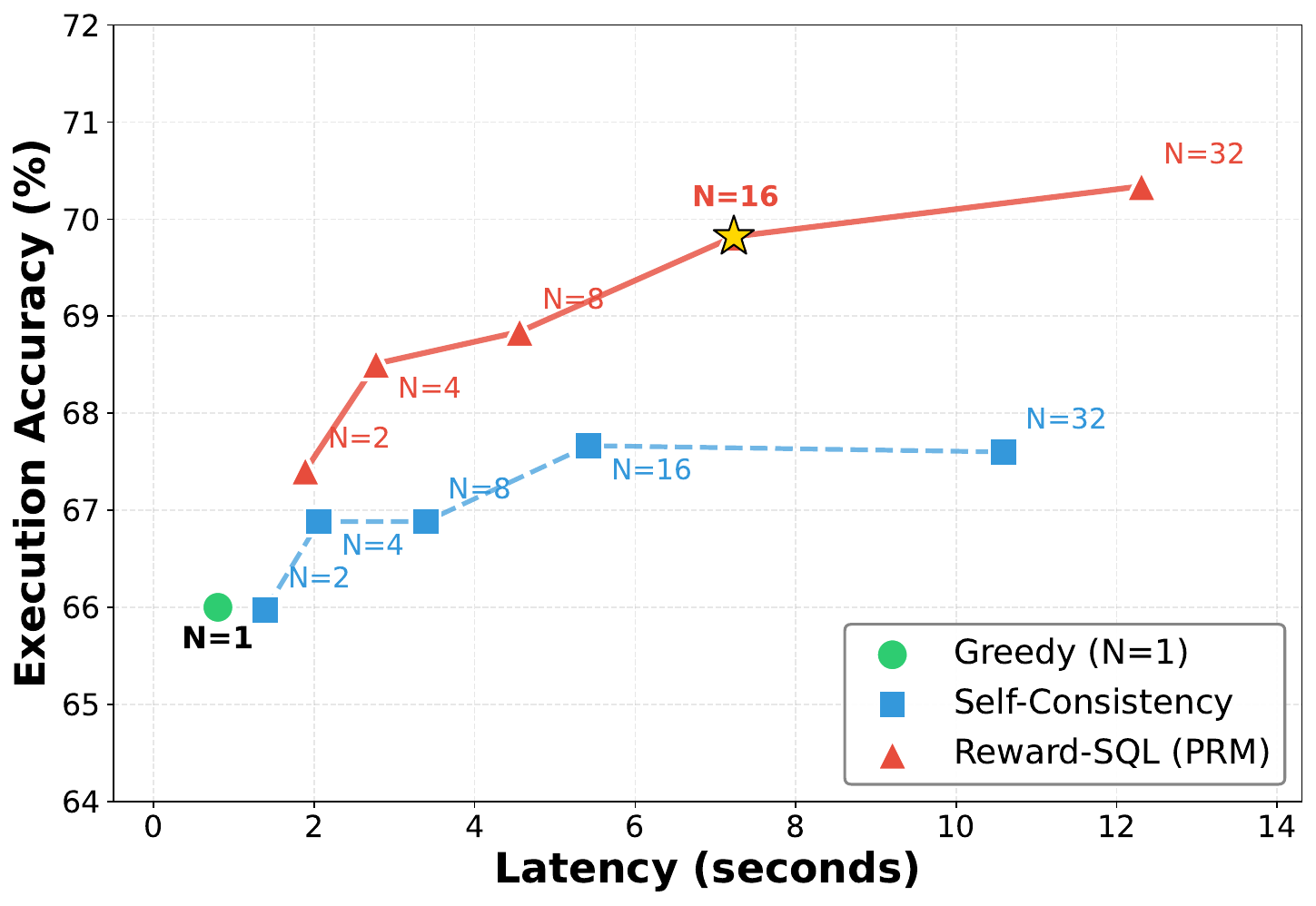}
\caption{\reviewtwo{Accuracy-Latency Pareto frontier on BIRD Dev., comparing greedy decoding, self-consistency (majority voting), and PRM-guided selection (ours).}}
\label{fig:pareto_frontier}
\end{figure}
\reviewtwo{
\stitle{Accuracy-Latency Trade-off.}
Figure~\ref{fig:pareto_frontier} presents the Pareto frontier comparing three inference strategies: (1) Direct SQL generation (\textbf{Greedy}), (2) \textbf{Self-Consistency} (majority voting over N samples), and (3)~\textbf{Reward-SQL} (PRM-guided selection, PRM@N).
While greedy decoding offers minimal latency (0.8s) with 66.0\% accuracy, scaling to PRM@16 achieves \textbf{69.6\% accuracy at 6.3s latency}, representing a substantial improvement of +3.6 percentage points. Self-consistency provides a middle ground but saturates quickly, even experiencing a slight decline of 0.1\% from N=16 to N=32. In contrast, PRM maintains steady improvements through effective candidate selection. Beyond N=16, however, PRM's marginal gains also diminish (only +0.7\% from N=16 to N=32) while latency doubles, establishing PRM@16 as the practical sweet-spot that balances accuracy gains against computational cost.
}

\begin{figure}[t]
\centering
\includegraphics[width=0.7\textwidth]{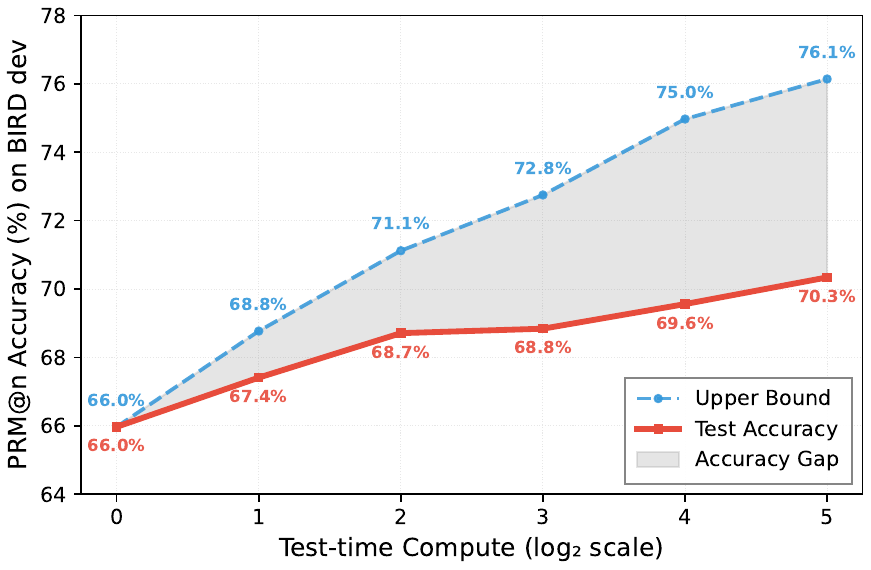}
\caption{Test-time compute vs accuracy on BIRD Dev. set. The upper bound (Pass@N) represents the theoretical maximum performance achievable with perfect selection, while PRM@N shows our model's actual performance.}
\label{fig:cost_tradeoff}
\end{figure}

\stitle{Scaling Behavior.} As shown in Figure~\ref{fig:cost_tradeoff}, performance scales logarithmically with the number of samples N. The upper bound (Pass@N) shows a nearly linear growth as test-time compute increases, climbing from 65.9\% at the lowest compute level to 76.1\% at the highest compute level. Our PRM@n accuracy follows a similar upward trajectory, starting at 65.9\% with only one sample and reaching 70.3\% with 32 samples. The gap between the theoretical upper bound and actual PRM@n performance gradually widens as compute increases, indicating untapped potential for further improvement through enhanced selection mechanisms.



\reviewone{\textbf{Query Execution Efficiency under CoCTEs Format.}
We evaluate the runtime overhead of CTE-based queries compared to flattened SQL on BIRD Dev. (SQLite). For fair comparison, we selected queries where \sys produced correct results and compared their latency against the corresponding gold SQL (optimized, flattened expert queries). Each query was executed 5 times with cache cleared between runs. On average, CoCTE queries execute in \textbf{230ms} versus \textbf{210ms} for gold SQL, representing a \textbf{<10\% overhead} (9.5\%). While this overhead is acceptable on BIRD-scale datasets, it may impact latency-sensitive enterprise applications with complex schemas. In future work we will explore (1) rule-based CTE transpilers to generate flattened SQL for production deployment, and (2) cost-aware reward modeling to optimize execution efficiency during training.}

\subsection{Error Analysis}
\label{subsec:error_analysis}

To understand how our approach improves SQL generation quality, we conduct a detailed error analysis comparing the baseline Qwen3-8B model with our full \sys framework at different test-time scaling budgets (PRM@8 and PRM@32).
We manually annotate error examples and categorize them according to a comprehensive error taxonomy.

Table~\ref{tab:error_analysis} presents a comprehensive breakdown of error categories.
Our analysis reveals several key insights:

\stitle{Major Improvements from Baseline to PRM@8:}
\begin{itemize}[leftmargin=*]
    \item \textbf{Hallucination Reduction (-6.00\%):} The most significant improvement is in reducing schema hallucinations, dropping from 6.26\% to 0.26\%. Process supervision effectively prevents the model from generating invalid schema elements by providing feedback when intermediate CTEs reference undefined database objects.
    \item \textbf{JOIN Key Accuracy (-2.34\%):} Structured reasoning through \cocte helps correctly identify foreign key relationships.
    \item \textbf{Schema Linking:} Significant reductions in Table Selection (-2.48\%) and Column Selection (-2.61\%) errors demonstrate improved schema comprehension.
    \item \textbf{Invalid SQL (-1.17\%):} The \cocte format naturally encourages syntactically valid SQL generation by decomposing complex queries into verifiable components.
\end{itemize}

\stitle{\reviewone{Gains from Scaling PRM@8 to PRM@32:}}
\reviewone{
Further increasing the sample budget from 8 to 32 yields additional error reductions, primarily in schema understanding:
\begin{itemize}[leftmargin=*]
    \item \textbf{Math Formula:} Notable 26\% relative improvement (46 to 34 instances), suggesting better handling of complex computations with increased sampling.
    \item \textbf{Table Selection:} Errors reduced by an additional 4.6\% (from 65 to 62 instances), indicating better table retrieval in complex multi-table scenarios.
    \item \textbf{Column Selection:} Errors reduced by an additional 5.9\% (from 51 to 48 instances), showing improved column mapping for nuanced queries.
\end{itemize}
}

\reviewone{
Overall, \sys reduces total errors from 45.57\% (baseline) to 31.49\% (PRM@8) and further to 29.66\% (PRM@32), representing a total relative improvement of 34.9\%.
The remaining challenges primarily involve semantic understanding (Filter Conditions, Domain Knowledge) where additional sampling provides limited benefit, suggesting these require stronger prior knowledge rather than increased test-time compute.}

\reviewthree{
\stitle{CTE-Level Error Analysis.}
To further validate that our \cocte decomposition does not introduce systematic errors, we manually analyzed 200 failed cases from the BIRD Dev. set, examining whether errors stem from CTE-level issues (e.g., wrong decomposition boundaries, redundant steps) or from underlying SQL generation problems.Table~\ref{tab:cte_error_breakdown} shows that 94.5\% of errors are not CTE-related, indicating that failures primarily originate from SQL generation challenges (e.g., incorrect filters, schema mismatches) rather than decomposition artifacts. Among CTE-specific issues, the most common is Data Loss in CTE Pipeline (4.0\%), where intermediate CTEs inadvertently filter out necessary rows through overly restrictive conditions. Cases of Redundant CTEs (1.0\%) and Wrong CTE Boundaries (0.5\%) are rare, demonstrating that our framework effectively learns appropriate decomposition granularity. 
\begin{table}[t]
\centering
\caption{\reviewthree{CTE-level error breakdown on 200 failed cases. The majority of errors are unrelated to CTE decomposition.}}
\label{tab:cte_error_breakdown}
\small
\begin{tabular}{lcc}
\toprule
\textbf{Error Category} & \textbf{Count} & \textbf{Percentage} \\
\midrule
Not CTE-related & 189 & 94.5\% \\
Data Loss in CTE Pipeline & 8 & 4.0\% \\
Redundant CTEs & 2 & 1.0\% \\
Wrong CTE Boundaries & 1 & 0.5\% \\
\midrule
\textbf{Total} & \textbf{200} & \textbf{100\%} \\
\bottomrule
\end{tabular}
\end{table}
}

%% file: secs/02related_work.tex
\section{Related Work}
\label{sec:related_work}

\textbf{SQL Decomposition and CoT.}
Natural language decomposition and finer-grained reasoning in Text-to-SQL can be broadly categorized into three directions. (i) Natural language decomposition. Representative works such as Reasoning-SQL \citep{pourreza2025reasoning-sql} and Omni-SQL \citep{li2025omnisql} first generate natural language reasoning traces—breaking the question into sub-goals and aligning them with SQL clauses—before producing the final query. Following this idea, Arctic-SQL \citep{yao2025arctic} and ExCoT\citep{zhai2025excot} integrate the reasoning process into the model using different RL algorithms. (ii) SQL structure decomposition. SQL-o1 \citep{lyu2025sql-o1} emphasizes structural planning: it decomposes a query into skeleton-level steps (e.g., deciding the overall layout and clause order) before filling in details. (iii) Operator-level decomposition. Alpha-SQL \citep{li2025alpha-sql} further refines this idea by decomposing queries into sequences of fine-grained operators (joins, filters, aggregations), treating SQL generation as a stepwise composition of primitive actions.

Despite these advances, most existing approaches decompose only at the natural language or abstract planning level, and the final SQL must still be assembled monolithically, which remains error-prone. A more robust form of SQL-level decomposition would require intermediate states that are executable and verifiable. In practice, Common Table Expressions (CTEs) serve exactly this role, allowing developers to structure complex queries into modular subqueries~\cite{airbyte_cte_guide,adv_sql_concepts_dataforge}. Recent works, such as ReFoRCE~\cite{deng2025reforce} have begun to incorporate CTEs for partial refinement, but this remains ad-hoc and limited. Overall, the CTE-based SQL decomposition proposed in this paper has not been systematically explored in Text-to-SQL, leaving a gap between natural language reasoning and executable query construction.

\stitle{RL-based Text-to-SQL Methods.}
Recent research has increasingly adopted RL to optimize Text-to-SQL models. On the training side, online RL approaches differ mainly in their reward design: SQL-R1 and Arctic-SQL employ execution accuracy (a lightweight but sparse supervision signal) as the sole reward in RL to improve the model's reasoning capability in Text-to-SQL~\cite{ma2025sql-r1,DBLP:journals/pvldb/LuoLFCT25}. To mitigate the sparse reward signals, Reasoning-SQL combines GRPO with partial execution-based rewards~\cite{pourreza2025reasoning-sql}. In parallel, offline preference optimization has gained prominence, where DPO-based \citep{Rafailov2023DPO} approaches align models via preference learning, and frameworks such as ExCoT integrate both off-policy and on-policy optimization~\cite{liu2025dpo-sql,zhai2025excot}.
On the inference side, the core idea is ``test-time scaling'', i.e., aggregating multiple candidates and then selecting the best one from them. The variants arise in the selection criteria, e.g., the self-consistency in C3-SQL~\cite{dong2023c3} the Best-of-N in STaR-SQL~\cite{he2025star-sql}, the Monte Carlo Tree Search (MCTS) in Alpha-SQL~\cite{li2025alpha-sql}, the self-rewarding in SQL-o1 \citep{lyu2025sql-o1}.

Reward design of the existing approaches is still dominated by outcome-based signals, i.e., whether the final SQL executes correctly. Such sparse feedback provides limited guidance during training and forces inference-time strategies to rely largely on frequency-based selection or heuristic search. To address this sparsity, several methods have explored process-level rewards. Execution-Guided Decoding (EGD) executes partial SQL during generation to prune invalid candidates~\cite{wang2018robust}. Reasoning-SQL introduces partial execution rewards to reduce sparsity during RL training~\cite{pourreza2025reasoning-sql}. Search-based methods such as Alpha-SQL (MCTS exploration) and SQL-o1 (self-rewarding search) also leverage process information to improve inference~\cite{li2025alpha-sql,lyu2025sql-o1}. ReEx-SQL \citep{DBLP:journals/corr/abs-2505-12768} integrates intermediate execution feedback but applies it to full-query revisions rather than stepwise decomposition into simpler executable sub-queries. 

Process Reward Models (PRMs) have demonstrated strong potential in other reasoning domains: in mathematics, they evaluate intermediate reasoning to address credit assignment issues~\cite{lightman2023let,wang2023math-shepherd}; in program synthesis, they assess partial code states to catch errors early~\cite{li2025codeprm, fan2026deepprep}.
However, existing Text-to-SQL methods remain largely heuristic or task-specific, lacking a learned and generalizable model to evaluate intermediate reasoning quality. Unlike mathematical or code reasoning, Text-to-SQL still lacks an explicitly trained PRM to systematically guide both learning and inference.
This gap motivates our work: we introduce a PRM-based framework that provides fine-grained, execution-aware supervision and principled search guidance for stepwise SQL generation.

%% file: secs/09conclusions.tex
\section{Conclusions}
\label{sec:conclusions}




In this paper, we have proposed \sys, a novel framework that addresses the key limitations of existing RL-based Text-to-SQL approaches through stepwise execution-aware reasoning and process-supervised rewards. We introduced \cocte, a divide-and-conquer reasoning format that decomposes complex SQL generation into a sequence of executable Common Table Expressions. To provide fine-grained supervision, we designed a process reward model that combines trajectory scoring with inverse entropy weighting, and integrated it into both GRPO training and Best-of-N inference, enabling stable optimization and principled trajectory selection. Extensive experiments on multiple benchmark datasets demonstrate the effectiveness of our method: \sys achieves 70.3\% execution accuracy on BIRD with an 8B model, outperforming baselines of comparable size and showing strong cross-domain generalization across five out-of-distribution benchmarks.